\documentclass[final]{cvpr}

\usepackage{graphicx}
\usepackage{amsmath,amssymb} % define this before the line numbering.
\usepackage{color}
\usepackage{microtype,booktabs,ragged2e,amsfonts,tabularx,url, courier,csquotes,dblfloatfix,amsmath, amsthm, amssymb, graphicx,amsfonts,multirow,longtable,subfig,algorithmic,algorithm,makecell} 

\usepackage{array}
\newcolumntype{L}[1]{>{\raggedright\let\newline\\\arraybackslash\hspace{0pt}}m{#1}}
\newcolumntype{C}[1]{>{\centering\let\newline\\\arraybackslash\hspace{0pt}}m{#1}}
\newcolumntype{R}[1]{>{\raggedleft\let\newline\\\arraybackslash\hspace{0pt}}m{#1}}
\usepackage{times}
\usepackage{epsfig}
\usepackage{graphicx}
\usepackage{amsmath}
\usepackage{amssymb}
\usepackage[pagebackref=true,breaklinks=true,colorlinks,bookmarks=false]{hyperref}

\def\cvprPaperID{7226} % *** Enter the CVPR Paper ID here
\def\confYear{CVPR 2021}

\begin{document}

%%%%%%%%% TITLE
\title{Prediction by Anticipation: An Action-Conditional Prediction Method based on Interaction Learning}

\author{Ershad Banijamali$^1$, Mohsen Rohani$^1$, Elmira Amirloo$^1$, Jun Luo$^1$, Pascal Poupart$^2$\\
$^1$Noah's Ark Laboratory, Huawei, Markham, Canada \\
$^2$School of Computer Science, University of Waterloo, Waterloo, Canada
 \\
{\tt\small  $\{$ershad.banijamali1,mohsen.rohani,elmira.amirloo,jun.luo1$\}$@huawei.com}, 
{\tt\small ppoupart@uwaterloo.ca}
}

\maketitle

%%%%%%%%% ABSTRACT
\begin{abstract}

In autonomous driving (AD), accurately predicting changes in the environment can effectively improve safety and comfort. Due to complex interactions among traffic participants, however, it is very hard to achieve accurate prediction for a long horizon. To address this challenge, we propose prediction by anticipation, which views interaction in terms of a latent probabilistic generative process wherein some vehicles move partly in response to the anticipated motion of other vehicles. Under this view, consecutive data frames can be factorized into sequential samples from an action-conditional distribution that effectively generalizes to a wider range of actions and driving situations. Our proposed prediction model, variational Bayesian in nature, is trained to maximize the evidence lower bound (ELBO) of the log-likelihood of this conditional distribution. Evaluations of our approach with prominent AD datasets NGSIM I-80 and Argoverse show significant improvement over current state-of-the-art in both accuracy and generalization.
\end{abstract}

\vspace{-.4cm}
\section{Introduction}
\vspace{-.2cm}

Predicting the future state of a scene with moving objects is a task that humans handle with ease. This is due to our understanding about the dynamics of the objects in the scene and the way they interact. Despite recent advancements in machine learning and especially deep learning, teaching machines such understanding remains challenging. 

In this paper, we are interested in solving the prediction task for multi-agent systems with interacting agents. 
We propose a new probabilistic approach for action-conditional prediction. The action-conditional prediction task is defined as finding the next observation from the scene given a sequence of observations, $\mathbf{o}_{1:t}$, and an action for an agent (ego agent), $\mathbf{a}_t$, i.e. optimizing $p(\mathbf{o}_{t+1}|\mathbf{o}_{1:t},\mathbf{a}_t)$. Our approach is based on the idea that sequences of data that include heavy interactions come from a probabilistic generative process wherein some reacting agents move partly in response to the \textit{anticipated} motion of some acting agents. In fact we split the observation into two sets of features: ego-features, $\mathbf{o}_t^{ego}$, which contains the features related to the ego agent, e.g. it's position, and environment-features, $\mathbf{o}_t^{env}$, which contains all other features than the ego-features including other agents or fixed objects in the scene. The action has an effect on both of these feature sets. In most of the previous works $p(\mathbf{o}_{t+1}|\mathbf{o}_{1:t},\mathbf{a}_t)$ is learned directly. This means predicting  ego- and  environment-features simultaneously, which is difficult due to interactions among the agents. However, the effect on the ego-features is usually much easier to model and often does not need to be learned.  Hence, we can decompose $p(\mathbf{o}_{t+1}|\mathbf{o}_{1:t},\mathbf{a}_t)$ into two steps: learning $p(\mathbf{o}^{ego}_{t+1}|\mathbf{o}_{1:t},\mathbf{a}_t)$ and then learning $p(\mathbf{o}^{env}_{t+1}|\mathbf{o}_{1:t},\mathbf{o}^{ego}_{t+1})$, i.e. we first learn how the action changes the ego-features and then learn how the environment reacts to this change.  We can often use domain knowledge to fix $p(\mathbf{o}^{ego}_{t+1}|\mathbf{o}_{1:t},\mathbf{a}_t)$ and then we can learn $p(\mathbf{o}^{env}_{t+1}|\mathbf{o}_{1:t},\mathbf{o}^{ego}_{t+1})$ from data. Learning $p(\mathbf{o}^{env}_{t+1}|\mathbf{o}_{1:t},\mathbf{o}^{ego}_{t+1})$ is much easier than learning $p(\mathbf{o}_{t+1}|\mathbf{o}_{1:t},\mathbf{a}_t)$ since we only have to learn to predict $\mathbf{o}^{env}_{t+1}$ based on the \textit{effect} of action $\mathbf{a}_t$ on the ego-features $\mathbf{o}^{ego}_{t+1}$.  Conditioning on $\mathbf{a}_t$ or $\mathbf{o}^{ego}_{t+1}$ is equivalent from an information theoretic perspective, but conditioning on $\mathbf{o}^{ego}_{t+1}$ allows the model to reason about interactions more easily since the ego effect is already anticipated.
Concretely, at each step of training, we apply the motion of an acting agent to the input observation, i.e. \textit{anticipating} the acting agent in its next position, and train a conditional density estimation model \cite{shu2017bottleneck} to predict where the other agents accordingly should be in the target observation, i.e. where they should move to \textit{partly in reaction to} the anticipated motion of the acting agent. 

In particular we consider the prediction task in the challenging framework of autonomous driving (AD), where there is a large number of agents in the scene and they have complicated interactions with each other. By thus recovering the latent generative process, our model is capable of achieving a higher capacity for prediction, i.e. handling a wider range of actions and driving situations.

% When the agents making predictions also act on the scene, such as in AD, the prediction problem becomes even harder. In such cases, \textit{action-conditional} prediction models become desirable, which predict how the dynamic objects, e.g. cars and pedestrians, would move in response to the actions of each other. 

To minimize preprocessing of data that can be costly and time-consuming at training and inference time, we adopt input and target representations in the form of occupancy grid maps (OGMs). The anticipating can be done through editing the original input OGM based on motion tracking of the acting vehicle. Working with OGMs also enables us to easily extend our model to predict the \textit{difference} between the input and target frames, rather than predict the target frame directly. We show that such difference learning works very well when the ego vehicle moves slowly, such as in dense urban traffic.  
% [JL: Very important technical detail but does not belong here in the intro. =>] We also transform the target frame such that it only differs from the imagined frame in the location of moving objects, as if it shows the reaction of these objects to the imagined position of the ego-vehicle. This ego-motion compensation, which is done using rule-based modules, in a substantial way reduces the high-dimensional action-conditional prediction problem to learning the interaction among the objects and expands the distribution of actions that can be predicted by the model. 
%We use a stochastic  module that works in the framework of variational autoencoders to learn the interactions. In fact, we propose a model that works almost similar to driving behavior of humans. Given current observation of the road and a history of the past observations, we as drivers decide to take an action. But, before immediately applying that action we \textit{imagine} how it will change our position in the road and \textit{predict} the reaction of other cars around us to this change\cite{bucchi2012traffic}.
Our contributions in this paper include:
\begin{itemize}
\itemsep-.3em 
    \item[\textbullet] A novel modular model for multi-step action-conditional prediction is proposed, which factorizes historical sequence data into samples drawn according to an underlying action-conditional distribution that covers a wider range of actions (including extreme actions) for predictions better than current state-of-the-art models. The method requires no labeling and is scalable with data.
    \item[\textbullet] An extension of the model for difference learning that outperforms the state-of-the-art prediction models in dense traffic.
    \item[\textbullet] Experiment results on two prominent AD datasets (NGSIM I-80 and Argoverse) with different interactions among vehicles, i.e. highway and urban area, demonstrating effective coverage of a wide variety of driving situations.
\end{itemize}

\section{Related Work}
A large body of literature on prediction tasks in AD is dedicated to prediction in low-dimensional space, i.e. position of the cars in the xy coordinates \cite{lee2017desire,deo2018convolutional,ma2019trafficpredict,casas2018intentnet, cui2019multimodal,bansal2018chauffeurnet,multiple2019Tang, salzmann2020trajectron++, huang2019stgat, park2018sequence, rhinehart2018r2p2, rhinehart2019precog, chai2019multipath, zeng2019end,liang2020pnpnet}. In most of these works the prediction task is done by finding the most probable paths for the objects in the environment using generative models. However, all of these methods need object detection and tracking (at least at training time), which is computationally expensive and requires labeled data. Moreover, any error in the object detection and tracking can affect the whole system and result in catastrophic failure. %Recently, in \cite{liang2020pnpnet} proposed an end-to-end method for trajectory prediction that takes the sensory data as input and performs objects detection and tracking in the inner loop 

Unsupervised prediction of OGMs \cite{elfes2013occupancy,tsardoulias2016review}  has also been studied recently. These methods do not model the effect of action in prediction and thus fail to capture the interactions. In \cite{milan2017online} and its extension \cite{dequaire2018deep} recurrent neural network (RNN)-based models were employed for OGM prediction. But, both models need data labeling and object detection. 
Authors in \cite{mohajerin2019multi} proposed a model for multi-step prediction of OGMs that produces state-of-the-art results on the KITTI dataset\cite{Geiger2013IJRR}. By \textit{removing} the ego-motion from the whole sequence, all the frames are mapped to a reference frame in which the ego-vehicle is frozen, i.e. the global location of the car is fixed. This way, only moving objects in the scene change their locations. There are two major differences between this work and ours. First of all, their predictions are not action-conditional.
% In best case, their model can be used for imitation learning, since we only have access to the drivers actions and the predicted OGMs are generated based on those actions. 
Secondly, since we \textit{compensate} the ego-motion step-by-step the global location of the ego car is not fixed.
%, i.e. the input and predicted outputs are in ego frame rather than a globally fixed frame. 
Therefore, in contrast to  \cite{mohajerin2019multi}, our model can predict for much longer horizons.

%In \cite{Engel2018} and \cite{Hoermann2017} Bayesian filtering is used to fuse different sensory data and build Dynamic OGMs (DOGMs). An occupancy state \cite{Nuss2018}, velocity and its uncertainty is associated with each cell. This information is then fed to a network which generates OGMs. The dataset used in both of these works are collected from a static sensor.
In \cite{Noguchi2012,Ohki2010,Gupta2008} OGM prediction is used for path planning. However, in \cite{Noguchi2012,Ohki2010} only one object type (human) is considered. \cite{Gupta2008} relies on object detection and the OGMs are updated using object models.
Model-predictive policy with uncertainty regularization (MPUR) \cite{henaff2019model}, is a state-of-the-art prediction and planning approach in this area. Although the model is successful in predicting the effect of existing actions in the training data, in the case of extreme actions it fails to predict a valid OGM. Alternatively, in \cite{bansal2018chauffeurnet} synthetic extreme actions are added to the training data in order to handle rare scenarios. However, since this is not a theoretically principled way for generalization, the performance of the algorithm is still limited by actions directly observed in the data. Moreover, addition of random actions not grounded in real interaction contexts can result in invalid scenarios that do not happen in real life, as the other cars do not react to the augmented actions. Finally, the method is an object tracking method with the aforementioned problems.

\section{Prediction by Anticipation}
The prediction task is described as follows. Given a set of $t$ observations from the scene, denoted by $\mathbf{o}_{1:t}$, and a set of $k$ actions denoted by $\mathbf{a}_{t:t+k-1}$, predict the future $k$ observations, $\mathbf{o}_{t+1:t+k}$. We try to solve this task by maximizing the conditional likelihood $p(\mathbf{o}_{t+1:t+k}| \mathbf{o}_{1:t}, \mathbf{a}_{t:t+k-1})$. The observations include (a) A bird's-eye view (BEV) image, denoted by $\mathbf{i}_t$ for time step $t$, in the form of an OGM with fixed position, e.g. in the middle of the image,  for the ego-vehicle. (b) Position and velocity of the ego-vehicle in each direction, which are denoted by $\mathbf{p}_t$ and $\mathbf{v}_t$, respectively, and are  referred to as  \textit{measurements}.
These two parts of the observation are both sensory data from the vehicle.
However,  we focus on predicting the images, as the position and velocity  can be deterministically computed given the actions, as described in the next sections. 
%This is why we assign them different names.
%%%%%%%%%%%%%%%%%%%%%%%%%%%%%%%%%%%%%%%%%%%%%%%%%%%%%%%%%
\begin{figure*}[!b]
    \centering
    \vspace{-.4cm}
    \includegraphics[trim = 0mm 5mm 0mm 5mm, width = 13.5cm]{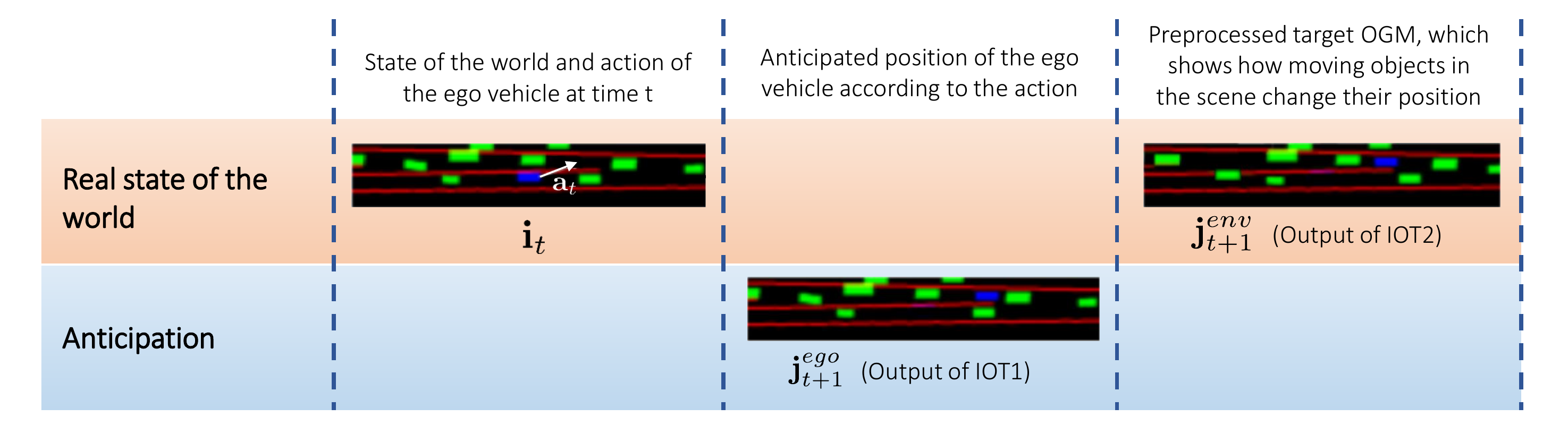}
    \vspace{-.2cm}
    \caption{\small Applying the effect of action on the OGMs: Left: OGM $\mathbf{i}_t$  and corresponding action at time $t$. Middle: the output of IOT1, $\mathbf{j}^{ego}_{t+1}$, after applying transformation on the ego-features of $\mathbf{i}_t$, . Right: the output of IOT2, $\mathbf{j}^{env}_{t+1}$, after applying transformation on $\mathbf{i}_{t+1}$.}
    \vspace{-.3cm}
    \label{fig: anticipation}
\end{figure*}
%%%%%%%%%%%%%%%%%%%%%%%%%%%%%%%%%%%%%%%%%%%%%%
%%%%%%%%%%%%%%%%%%%%%%%%%%%%%%%%%%%%
\subsection{Base model}
Let's assume $\mathbf{o}_t = \{\mathbf{o}_t^{ego}, \mathbf{o}_t^{env} \}$,   where $\mathbf{o}_t^{ego} $ includes the features related to the ego-agent (ego-vehicle), i.e. $\mathbf{p}_t$, $\mathbf{v}_t$, and parts of the OGM  related to the ego-vehicle, denoted by $\mathbf{i}_t^{ego}$. Therefore $\mathbf{o}_t^{ego} = \{\mathbf{p}_t, \mathbf{v_t}, \mathbf{i}_t^{ego} \}$. The environment-features,  $\mathbf{o}_t^{env}$, include all other features in the observation than the ego-features. In our application this includes parts of the OGM that represent other objects in the scene, e.g. other agents, maps, static objects, etc. Therefore, $\mathbf{o}^{env}_{t} = \{ \mathbf{i}_t^{env}\}$. The actions of the ego-vehicle (ego-actions) affect both ego- and environment-features. The first-order effect of the action is on the ego-features, which can be determined using our \textit{prior knowledge} about the dynamics of the ego-vehicle. The second-order effect is on the environment-feature that should be learned from data. 
Thus, to maximize the conditional log-likehihood of $\log p(\mathbf{o}_{t+1:t+k}| \mathbf{o}_{1:t}, \mathbf{a}_{t:t+k-1})$, we first split it into $k $ autoregressive steps where at each time step the previous prediction is fed-back to the model. Then for each time step we factorize the log-likelihood into two terms:
\begin{align}
\vspace{-.4cm}
     \nonumber
     \log p(\mathbf{o}_{t+1}| \mathbf{o}_{1:t}, \mathbf{a}_{t}) & =  \log p(\mathbf{o}^{ego}_{t+1}| \mathbf{o}_{t}, \mathbf{a}_{t}) \\ &+ \log p(\mathbf{o}^{env}_{t+1}| \mathbf{o}_{1:t}, \mathbf{o}^{ego}_{t+1}). 
     \label{eq: first_like}
     \vspace{-.4cm}
\end{align}
Note that to determine the next ego-features, we only need the current observation  and  action. For the second term, we assume $p(\mathbf{o}^{env}_{t+1}| \mathbf{o}_{1:t}, \mathbf{o}^{ego}_{t+1}, \mathbf{a}_{t}) = p(\mathbf{o}^{env}_{t+1}| \mathbf{o}_{1:t}, \mathbf{o}^{ego}_{t+1})$.  This assumption is based on the fact that the environment does not observe $\mathbf{a}_t$ directly. In other words, the other agents only observe the \textit{effect} of  the ego-action and not the actual action itself. Also, it is worth mentioning that learning $p(\mathbf{o}^{env}_{t+1}|  \mathbf{o}_{1:t}, \mathbf{o}^{ego}_{t+1})$ does not mean that we assume causality between $\mathbf{o}^{env}_{t+1}$  and $\mathbf{o}^{ego}_{t+1}$. We simply learn a distribution where there is a non-causal correlation between $\mathbf{o}^{env}_{t+1}$  and $\mathbf{o}^{ego}_{t+1}$. The idea is that other agents anticipate  $\mathbf{o}^{ego}_{t+1}$ and act accordingly. The conditional distribution $p(\mathbf{o}^{env}_{t+1}| \mathbf{o}_{1:t}, \mathbf{o}^{ego}_{t+1})$ models the noise in this anticipation. But since there is a strong correlation between $\mathbf{o}^{env}_{t+1}$  and $\mathbf{o}^{ego}_{t+1}$, then conditioning on  $\mathbf{o}^{ego}_{t+1}$ will be very informative. 

To implement this idea, we design a modular model that consists of rule-based and learning-based modules. The rule-based modules implement the deterministic parts of Eq. \ref{eq: first_like}, which are based on our prior knowledge and geometry of the problem. The learning-based module, which we call the prediction module, performs the prediction task by learning the interactions among the agents.  
%We employ a special case of the conditional variational Bayes to efficiently estimate the parameters of the prediction module. 
%%%%%%%%%%%%%%%%%%%%%%%%%%%%%%%%%%%%%%%%
\vspace{-.2cm}
\subsubsection{Rule-based modules}
\vspace{-.1cm}
These modules are responsible for factorization according to the presumed underlying generative process. Both parts of the observations are changed to account for the effect of the action. This is done using deterministic functions for measurements and deterministic transformations (rotation and translation) for the OGMs. 

\begin{figure}[!t]
  \vspace{-.35cm}
    \centering
    \includegraphics[trim=10mm 10mm 10mm 00mm, width = 6cm]{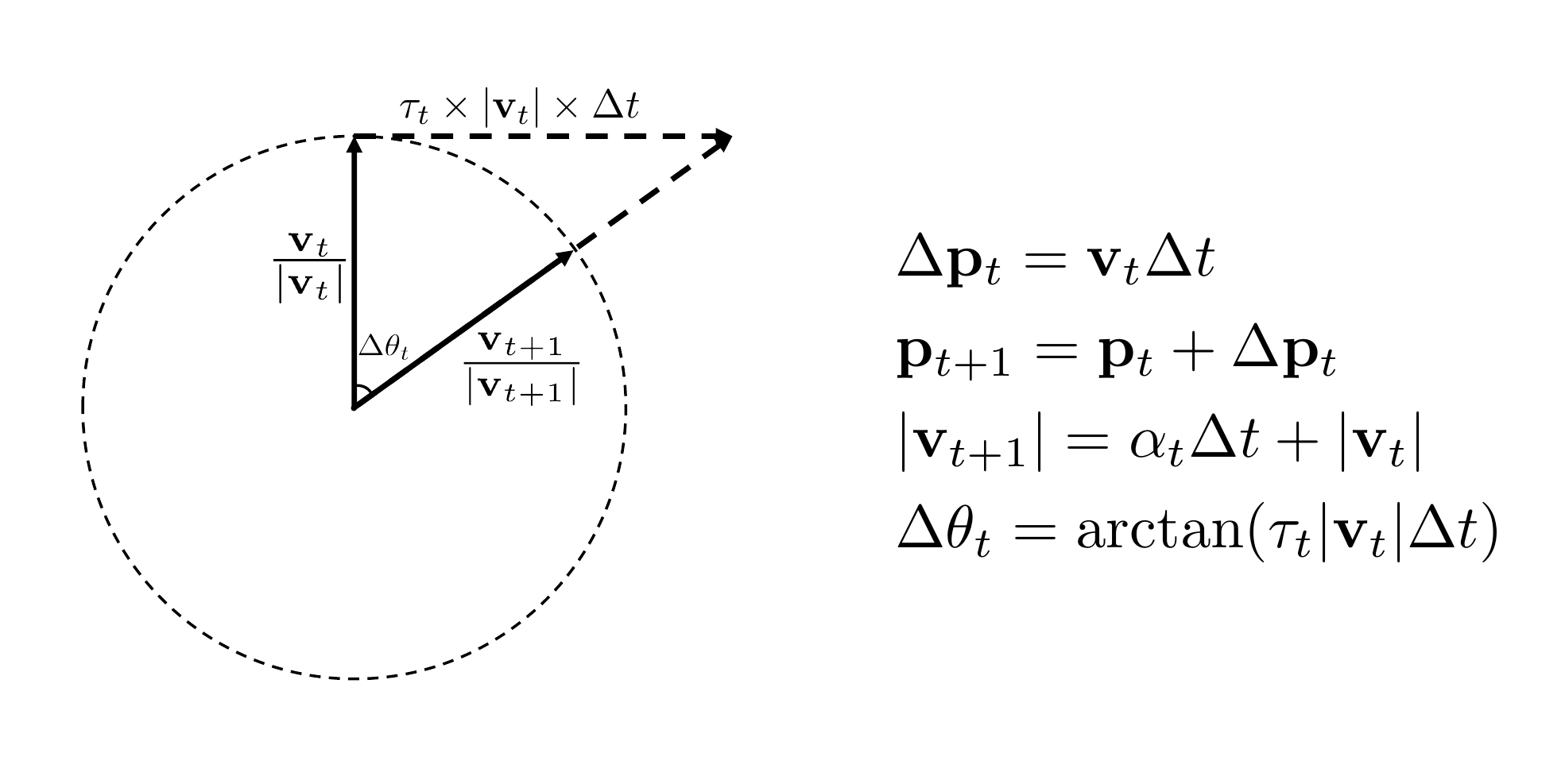}
    \vspace{-.3cm}
    \caption{\small Computing the effect of actions on the future position, velocity and change in the direction of the car.% $\Delta t$ depends on sampling frequency. We consider the sampling rate of $10$, therefore $\Delta t = 0.1$.
    }
    \vspace{-.5cm}
    \label{fig: measure}
\end{figure}

%%%%%%%%%%%%%%%%%%%%%%%%%%%%%%%%%%%%%%%
\noindent \textbf{Measurements estimator module:} Given the actions and measurements at each time step, this module computes the next measurements according to the update rule shown in Fig.  \ref{fig: measure}. Actions are two-dimensional, which include acceleration, $\alpha$, and rotation of the steering wheel, $\tau$, $\mathbf{a}_t = [ \alpha_t,\tau_t$]. This module also provides elements of translation and rotation matrices for the image processing modules:
\begin{equation}
\vspace{-.1cm}
        \mathbf{p}_{t+1}, \mathbf{v}_{t+1}, \Delta \mathbf{p}_t, \Delta \theta_t = f_m (\mathbf{p}_t,\mathbf{v}_t,\mathbf{a}_t).
        \label{eq: fm}
        \vspace{-.1cm}
\end{equation}
%
%\begin{align}
    %\Delta \mathbf{p}_t &=\mathbf{v}_t \Delta t, \hspace{1cm} \mathbf{p}_{t+1} = \mathbf{p}_t + \Delta \mathbf{p}_t \\ \nonumber
 %   |\mathbf{v}_{t+1}| &= \alpha_t  \Delta t+ |\mathbf{v}_t| \\ \nonumber
%    \Delta \theta_t &= \arctan (\tau_t \times |\mathbf{v}_t| \times \Delta t ),
%\end{align}
%%%%%%%%%%%%%%%%%%%%%%%%%%%%%%%%%%%%%%%%%%%%%%%%%%%
\begin{figure*}[!t]
    \centering
    \includegraphics[width = 13cm]{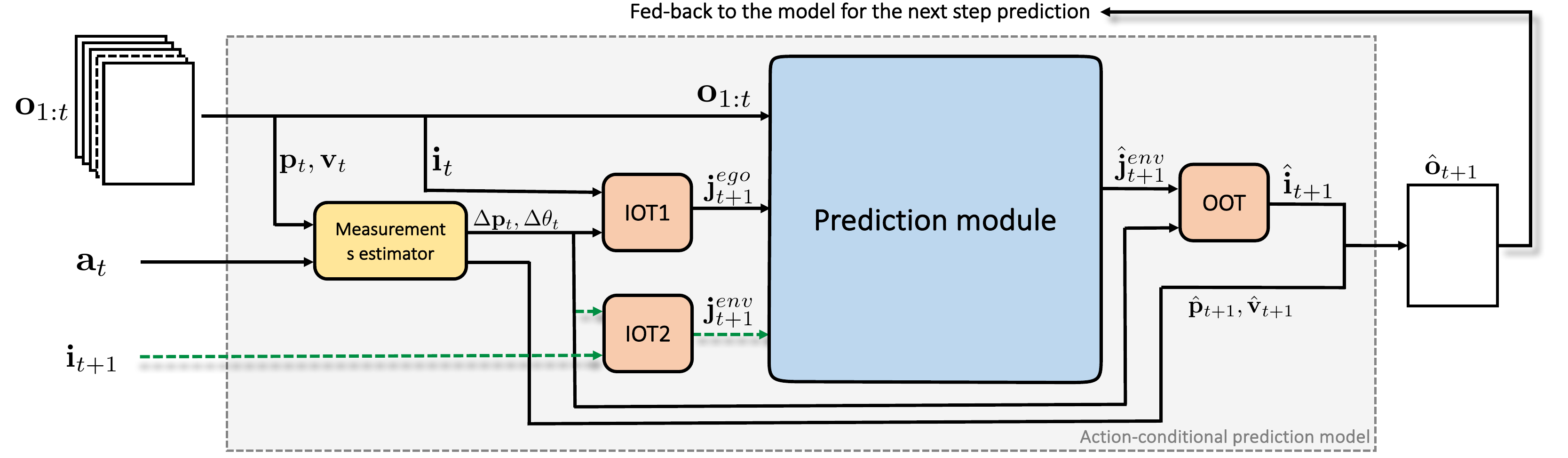}
    \vspace{-.4cm}
    \caption{\small The prediction model with all of its components. The measurement estimation module updates the position and velocity of the ego-vehicle and provides transformation parameters for OGM transformer modules. IOT1 and IOT2 provide information about the first and second-order effect of action on the OGM, respectively. Prediction module is trained to minimize the cost in Eq. \ref{eq: objective}. 
    \vspace{-.6cm}
    %The random samples in the latent space are from the variational approximation distribution $q(\mathbf{z}_t|\mathbf{o}_{1:t}, \mathbf{j}^{ego}_{t+1}, \mathbf{i}^{\text{all}}_{t+1})$ at the training time and from the conditional prior $p(\mathbf{z}_t|\mathbf{o}_{1:t}, \mathbf{j}^{ego}_{t+1})$ at the test time.
    }
    \label{fig:model}
\end{figure*}
%%%%%%%%%%%%%%%%%%%%%%%%%%%%%%%%%%%%%%%%%%%%%%%%%%%
\noindent\textbf{Input OGM transformation modules (IOTs):} In order to apply the first-order effect of the action in the OGM, we change the position of the ego-vehicle in the current OGM, $\mathbf{i}_t$. This transformation takes the ego-vehicle to its \textit{anticipated} position at time $t+1$ based on the action. We denote the module that performs this transformation by IOT1, and  the output is denoted by $\mathbf{j}^{ego}_{t+1}$:
\begin{equation}
\vspace{-.1cm}
    \mathbf{j}^{ego}_{t+1} = \text{IOT1}(\mathbf{i}_{t}, \Delta \mathbf{p}_t, \Delta{\theta}_t).
    \vspace{-.1cm}
\end{equation}
Moreover, at each time step of training we preprocess the target OGM, $\mathbf{i}_{t+1}$, to account for the second-order effect of the action. That is, we transform the whole target OGM in a way that the ego-vehicle has the same position as in $\mathbf{j}^{ego}_{t+1} $ . We denote the module for this transformation and its output by IOT2 and $\mathbf{j}^{env}_{t+1}$, accordingly.
\begin{equation}
\vspace{-.1cm}
    \mathbf{j}^{env}_{t+1} = \text{IOT2}(\mathbf{i}_{t+1}, \Delta \mathbf{p}_t, \Delta{\theta}_t)
    \vspace{-.1cm}
\end{equation}
Fig. \ref{fig: anticipation} shows the output of these two modules for a sequence of current and target OGMs. Note that after these two transformations, only the position of the moving objects will be different in $\mathbf{j}^{ego}_{t+1}$ and $\mathbf{j}^{env}_{t+1}$, while map information and structure of the fixed objects in the OGM, e.g. buildings, tree, parked cars, etc.,  will be the same. Both $\mathbf{j}^{ego}_{t+1}$ and $\mathbf{j}^{env}_{t+1}$ are fed to the prediction module during training.
%and $\mathbf{j}^{env}_{t+1}$ is the target to be predicted. 

We should clarify here that $\mathbf{j}_{t+1}^{ego}$ and $\mathbf{j}^{env}_{t+1}$ notations are used to emphasize the change after applying the motion of the ego-vehicle.  More generally, we will use $\mathbf{j}$ to denote OGMs resulting from deterministic transformations and $\mathbf{i}$ to denote the ego-centred OGMs. In fact, our original goal of optimizing for the stochastic mapping $p(\mathbf{i}_{t+1}^{env}|\mathbf{o_{1:t}}, \mathbf{i}_{t+1}^{ego})$ is equivalent to optimizing for $p(\mathbf{j}_{t+1}^{env}|\mathbf{o_{1:t}}, \mathbf{j}_{t+1}^{ego})$, up to some deterministic transformations.
\noindent\textbf{Output OGM transformation module (OOT):}
This module takes the predicted frame and transforms the whole frame  using $\Delta \theta_t$, and $\Delta \mathbf{p}_t$ such that the ego-vehicle goes back to its fixed position in $\mathbf{i}$ OGMs and environment-features change accordingly. 
%In fact, OOT's function is the inverse of IOT2's. 
The output should ideally be $\mathbf{i}_{t+1}$, which is fed back to the model for the next step prediction. This module is necessary to close the loop for \textit{multi-step} prediction.
%
%%%%%%%%%%%%%%%%%%%%%%%%%%%%%%%%%%%%%%%%%%%%
\vspace{-.3cm}
\subsubsection{Learning-based module: Prediction module}
\vspace{-.1cm}
The prediction module is the core of our model that predicts how the  environment partially reacts to the ego-action, i.e. learns $p(\mathbf{j}^{env}_{t+1}|\mathbf{o}_{1:t}, \mathbf{j}^{ego}_{t+1})$. Using the IOT1 and IOT2 modules the geometry of the target frame, $\mathbf{j}^{env}_{t+1}$, remains the same as the input frame, $\mathbf{j}^{ego}_{t+1}$, at each time step, regardless of the ego-action. Therefore $p(\mathbf{j}^{env}_{t+1}|\mathbf{o}_{1:t}, \mathbf{j}^{ego}_{t+1})$ is a smoother function than the original objective function, $ p(\mathbf{o}_{t+1}|\mathbf{o}_{1:t}, \mathbf{a}_t)$. Thus, $p(\mathbf{j}^{env}_{t+1}|\mathbf{o}_{1:t}, \mathbf{j}^{ego}_{t+1})$  is intuitively easier to learn.
%This is especially important for the case of self-driving cars where the prediction model should reliably predict the future images online. 
% The prediction module only needs to predict the location of the dynamic objects and rotations and translations caused by the ego-action do not need to be learned during the training of the model.  
Despite this simplification, the two objectives have the same optimum point, i.e. maximizing one leads to maximizing the other.
The first term in Eq. \ref{eq: first_like} is deterministic and can be removed from the optimization. Also, $\mathbf{j}^{env}_{t+1}$ is uniquely determined by the action $\mathbf{a}_t$ (given $\mathbf{i}_{t+1}$). Consequently, we can re-write the second term of Eq. \ref{eq: first_like} as $\log p(\mathbf{j}^{env}_{t+1}|\mathbf{o}_{1:t}, \mathbf{j}^{ego}_{t+1})$ . 
%Since the image and measurements in $\mathbf{o}_{t+1}$ are conditionally independent, given the past observations and the current action, we can write the log-likelihood in the following form:
%
% \begin{align}
%          \log p(\mathbf{o}_{t+1}|\mathbf{o}_{1:t}, \mathbf{a}_t) = \log p(\mathbf{i}_{t+1}|\mathbf{o}_{1:t}, \mathbf{a}_t)  + \log p(\mathbf{p}_{t+1}, \mathbf{v}_{t+1} |\mathbf{o}_{1:t}, \mathbf{a}_t),
%         \label{eq: like1}
% \end{align}
% %
% According to Eq. \ref{eq: fm}: $p(\mathbf{p}_{t+1}, \mathbf{v}_{t+1} |\mathbf{o}_{1:t}, \mathbf{a}_t) =  p(\mathbf{p}_{t+1}, \mathbf{v}_{t+1} |\mathbf{p}_{t}, \mathbf{v}_{t}, \mathbf{a}_t)
%       = \delta(f_m (\mathbf{p}_{t}, \mathbf{v}_{t}, \mathbf{a}_t)), $  
% where $\delta(.)$ is the Dirac delta function. Therefore the second likelihood in Eq. \ref{eq: like1} can be removed. Also, both $\mathbf{j}^{ego}_{t+1}$ and $\mathbf{i}^{\text{all}}_{t+1}$ are one-to-one functions of the action $\mathbf{a}_t$ (given $\mathbf{i}_t$ and $\mathbf{i}_{t+1}$). Consequently, we can re-write the first term of Eq. \ref{eq: like1} as $\log p(\mathbf{i}^{\text{all}}_{t+1}|\mathbf{o}_{1:t}, \mathbf{j}^{ego}_{t+1})$ .

\noindent \textbf{Bottleneck conditional density estimation: }
We maximize the conditional log-likelihood $\log p( \mathbf{j}^{env}_{t+1}| \mathbf{o}_{1:t},  \mathbf{j}^{ego}_{t+1})$ in the framework of variational Bayes. We build our model upon a Bottleneck Conditional Density Estimation (BCDE) \cite{shu2017bottleneck} model, a special variant of the conditional variational autoencoders (CVAEs) \cite{sohn2015learning}. 
The latent code in BCDE acts as bottleneck of information and not just a source of randomness. The prior on the latent variable in BCDE is conditioned on the input. Such conditioning makes the model less prone to overfitting as it allows learning the distribution of the latent code conditioned on input, which is especially helpful for prediction with large horizon. 
 %%%%%%%%%%%%%%%%%%%%%%%%%%%%%%%%%%
%
\begin{figure}[!t]
    \centering
    \includegraphics[width=8cm]{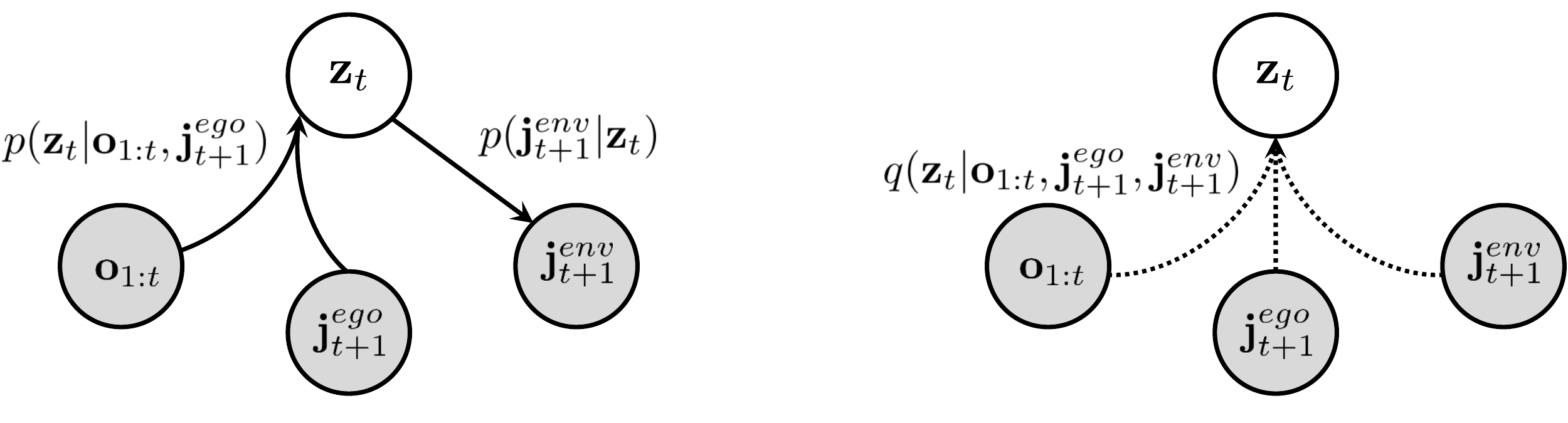}
        \vspace{-.4cm}
    \caption{\small Graphical model at time $t$: Left: Generative links, $p(.)$. Right: Variational links, $q(.)$. Observable variables are gray.}
    \vspace{-.2cm}
    \label{fig:gm}
    \vspace{-.4cm}
\end{figure} 
%
%%%%%%%%%%%%%%%%%%%%%%%%%%%%%%%%%%
 We consider the graphical model in Fig. \ref{fig:gm} at each time step for this prediction task. 
According to our definition of the approximating variational distribution in the graphical model, and also considering $\mathbf{z}_t$ as an information bottleneck between the input, $\mathbf{o}_{1:t}$ and $\mathbf{j}^{ego}_{t+1}$, and the target, $\mathbf{j}^{env}_{t+1}$, the ELBO to be maximized will have the following form:
\vspace{-.1cm}
\begin{align}
    \nonumber
    \log p(\mathbf{j}^{env}_{t+1}|\mathbf{o}_{1:t}, \mathbf{j}^{ego}_{t+1}) &\geq \mathbb{E}_{q^*(\mathbf{z}_t)} [\log p(\mathbf{j}^{env}_{t+1}|\mathbf{z}_t)] \\ &- \text{KL} \big (q^*(\mathbf{z}_t) || p(\mathbf{z}_t|\mathbf{o}_{1:t}, \mathbf{j}^{ego}_{t+1} ) \big ),
    \label{eq: lowerbound}
\end{align}
where $q^*(\mathbf{z}_t) = q(\mathbf{z}_t|\mathbf{o}_{1:t}, \mathbf{j}^{ego}_{t+1}, \mathbf{j}^{env}_{t+1} )$. We implement each of the conditional probability distributions in Eq. \ref{eq: lowerbound} using a neural network and denote the parameters of $p_{\psi}(.)$ and $ q_{\phi}(.)$ by $\psi$ and $\phi$, respectively. 
%%%%%%%%%%%%%%%%%%%%%%%%%%%%%%%%%%%%%%%%%%%%%%%%%%%
\begin{figure*}[!t]
    \centering
    \includegraphics[width = 14cm]{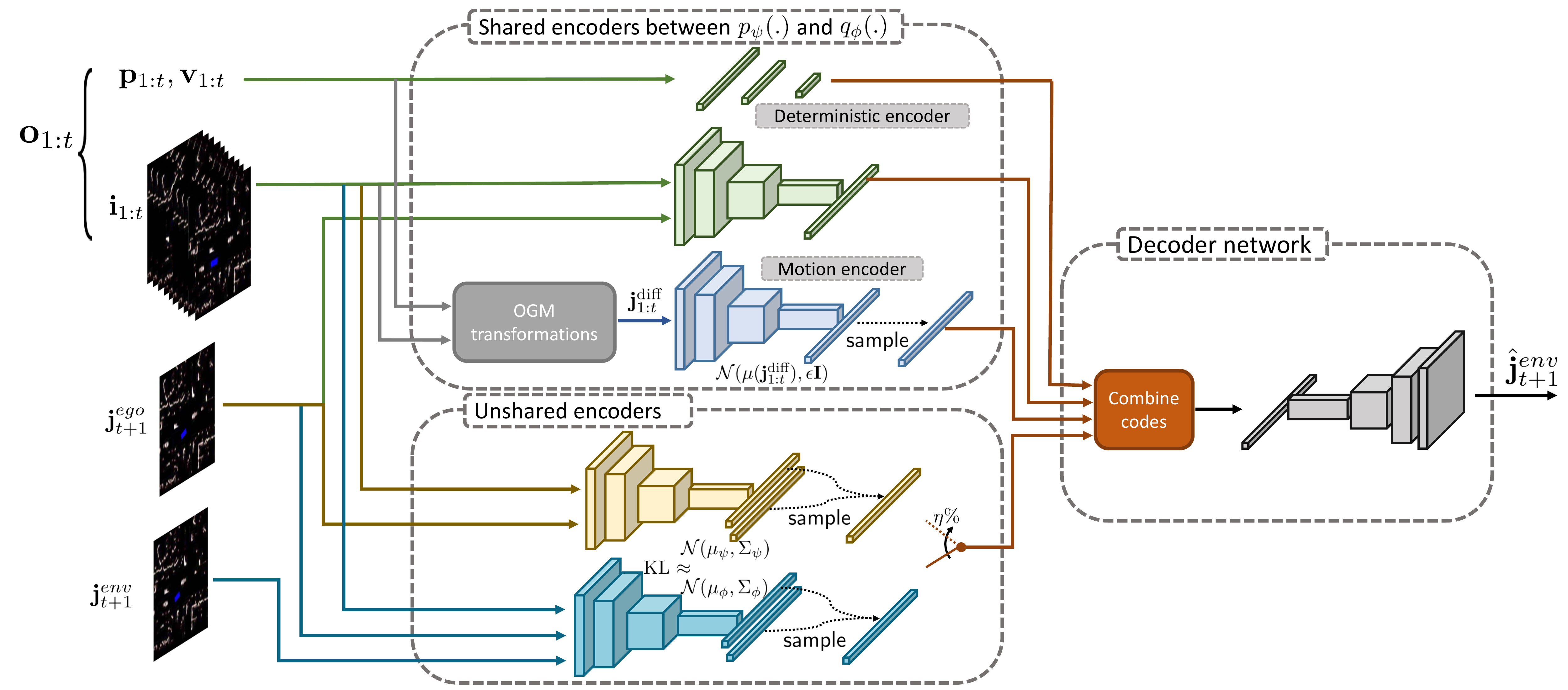}
    \vspace{-.4cm}
    \caption{\small The prediction module at the training time. The measurements are encoded using fully-connected networks, while we use convolutional neural networks to encode and decode OGMs.  
    }
    \vspace{-.6cm}
    \label{fig:pm_h}
\end{figure*}
%%%%%%%%%%%%%%%%%%%%%%%%%%%%%%%%%%%%%%%%%%%%%%%%%%%

%%%%%%%%%%%%%%%%%%%%%%%%%%%%%%%%%%%%%%%%%%
\noindent \textbf{Reconstruction loss and structural similarity: }The first term in the ELBO in Eq. \ref{eq: lowerbound}, can be interpreted as a reconstruction loss in the pixel space that measures the difference between the target OGM, $\mathbf{j}^{env}_{t+1}$, and predicted OGM, $\hat{\mathbf{j}}^{env}_{t+1}$, and we denote it by $\mathcal{D}(\mathbf{j}^{env}_{t+1} , \hat{\mathbf{j}}^{env}_{t+1})$. Depending on the type of values in the OGM being continuous or binary we consider Gaussian (with identity covariance matrix) or Bernoulli distributions for the output and replace $\mathcal{D}(\mathbf{j}^{env}_{t+1} , \hat{\mathbf{j}}^{env}_{t+1})$ with the mean squared error (MSE) or the cross entropy (CE), respectively. We also add an auxiliary term to our reconstruction loss that computes the Structural Similarity Index (SSIM) \cite{wang2004image} loss between prediction and target. The experiments show the effectiveness of adding this term in improving the quality of the predicted frames. The weight of the SSIM term, $\lambda$, is set using the validation set. 
%We provide an ablation study on this term in the supplementary material.
%%%%%%%%%%%%%%%%%%%%%%%%%%%%%%%%%%%%%%%%%%%%%%%%%%

\noindent \textbf{Code splitting and sampling from prior: }
The second term, is a KL divergence regularization that minimizes the distance between the output distributions of $p_{\psi}(.)$ and $q_{\phi}(.) $ encoders. Since the observations are highly dynamic with many objects in the scene, merely minimizing the KL divergence does not provide a proper training for the $p_{\psi}(.) $ encoder. Therefore we employ two ideas to better match these two distributions. 
%encoder of matching the output distributions of p and q  to capture different aspects of the input data 

1) We split the latent code $\mathbf{z}_t$ into two parts with two different sets of encoders. For the first part we do not use the target frame as the input of the $q_{\phi}(.) $ encoder and therefore its parameters can be shared with the $p_{\psi}(.)$ encoder, which guarantees the minimization of KL divergence. We call this part, \textit{shared code} and since it only encodes the previous and current observations we assume it is partly deterministic.
%, a deterministic part, a stochastic part, and a motion encoding part (described in the next section). The deterministic part encodes the input of the model, $\mathbf{o}_{1:t}$ and $\mathbf{j}_{t+1}^{ego}$, and KL divergence for this part is minimized by sharing the parameters between the encoder of $p_{\psi}(.)$ and $q_{\phi}(.)$. 
For the second part, called \textit{unshared code}, we assume Gaussian distributions for both the conditional prior $p_{\psi}(\mathbf{z}_t|\mathbf{o}_{1:t}, \mathbf{j}^{ego}_{t+1} )$ and the variational posterior $q_{\phi}(\mathbf{z}_t|\mathbf{o}_{1:t}, \mathbf{j}^{ego}_{t+1}, \mathbf{j}^{env}_{t+1} )$ and minimize their KL divergence. This part of $\mathbf{z}_t$ encodes information about the target, $\mathbf{j}^{env}_{t+1}$, and its stochasticity represents the uncertainty about the future. By combining this splitting idea with the BCDE model we introduce another important difference with the vanilla CVAE model, i.e. instead of concatenating the stochastic code with our high-dimensional input, we concatenate it with an encoded version of the input  that keeps only useful information. This makes the structure of the decoder simpler with far fewer  parameters, and therefore easier to train. 

2) To make sure that the encoder of $p_{\psi}(.)$ for the unshared code is trained properly, we randomly switch between the stochastic samples of the $p_{\psi}(.)$ and $q_{\phi}(.)$ encoders, i.e. $\eta \%$ of the time the samples are drawn from $p_{\psi}(\mathbf{z}_t|\mathbf{o}_{1:t}, \mathbf{j}^{ego}_{t+1} )$ instead of $q_{\phi}(\mathbf{z}_t|\mathbf{o}_{1:t}, \mathbf{j}^{ego}_{t+1}, \mathbf{j}^{env}_{t+1} )$. This way the $p_{\psi}(.)$ encoder is  trained by backpropagating both errors of the KL term and  the reconstruction term. In our experiments we set $\eta = 10$.
The final training objective to be minimized is:
\vspace{-.1cm}
\begin{align}
    \nonumber
     \mathcal{L}_t &=  
     \underbrace{\mathcal{D}(\mathbf{j}^{env}_{t+1} , \hat{\mathbf{j}}^{env}_{t+1}) \!+ \!\lambda \big( 1- \text{SSIM} (\mathbf{j}^{env}_{t+1},\hat{\mathbf{j}}^{env}_{t+1}) \big)}_{\mathcal{L}_t^{\text{rec.}}}  \\ &\!+\! \underbrace{\text{KL} \big (q_{\phi}(\mathbf{z}_t|\mathbf{o}_{1:t}, \mathbf{j}^{ego}_{t+1}, \mathbf{j}^{env}_{t+1} ) || p_{\psi}(\mathbf{z}_t|\mathbf{o}_{1:t}, \mathbf{j}^{ego}_{t+1} ) \big )}_{\mathcal{L}_t^{\text{KL}}}. 
    \label{eq: objective}
\end{align}
For a multi-step prediction with horizon $k$ a summation over $\mathcal{L}_t$ is minimized: $   \min \limits_{\psi, \phi} \sum_{j=0}^{k-1} \mathcal{L}_{t+j}$.
%%%%%%%%%%%%%%%%%%%%%%%%%%%%%%%%%%%%%%%%%%%%%%%%
\subsection{Difference Learning (DL)}
%%%%%%%%%%%%%%%%%%%%%%%%%%%%%%%%%%%%%%%%%%%%
\begin{figure}[!h]
    \centering
    \vspace{-.3cm}
    \includegraphics[width = 7cm]{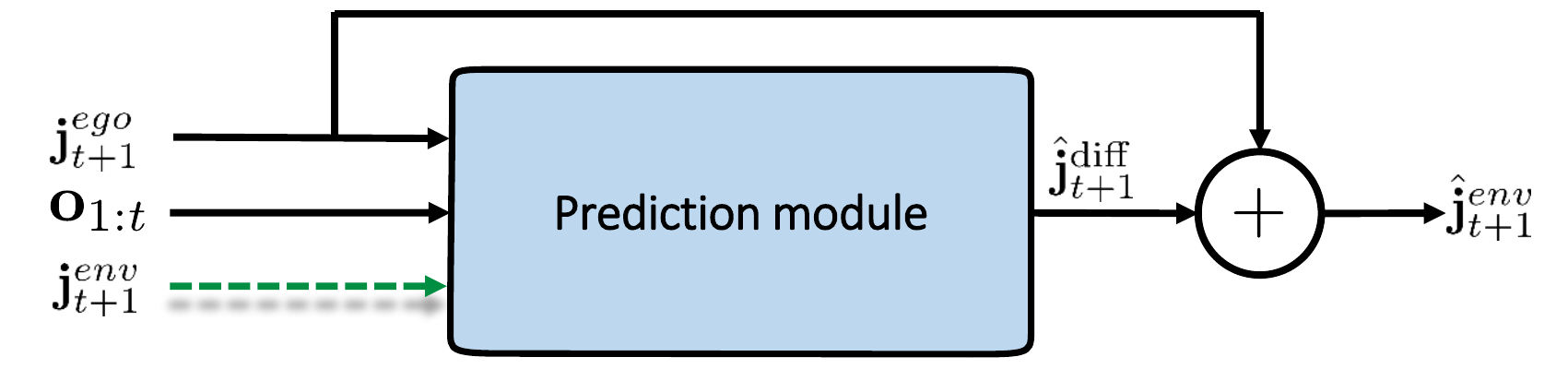}
    \vspace{-.2cm}
    \caption{\small Difference learning module}
    \vspace{-.3cm}
    \label{fig:dm}
\end{figure}
%%%%%%%%%%%%%%%%%%%%%%%%%%%%%%%%%%%%%%%%%%%%
%
After the OGM transformations, the difference between $\mathbf{j}^{ego}_{t+1}$ and $\mathbf{j}^{env}_{t+1}$ is only in the position of the moving objects. Therefore we also propose a variant of our model that explicitly learns the difference between the two OGMs, denoted by $\mathbf{j}^{\text{diff}}_{t+1} = \mathbf{j}^{env}_{t+1} - \mathbf{j}^{ego}_{t+1}$. In fact,  $\mathbf{j}^{\text{diff}}_{t+1}$ represents the motion of other agents in the pixel space. 
%For low-speed dense traffic, the difference between two consecutive frames, $\mathbf{j}^{ego}_{t+1}$ and $\mathbf{j}^{env}_{t+1}$, is small. In these cases predicting the whole frame of $\mathbf{j}^{env}_{t+1}$ is a harder task than predicting just the difference between the two frames. This is essentially due to the fact that according to the reconstruction loss in Eq. \ref{eq: objective}, the model tends to learn the static objects in the frame and uses very limited power to learn the movements. Therefore, for such scenarios, we use an extension of our model that explicitly learns the difference between $\mathbf{j}^{ego}_{t+1}$ and $\mathbf{j}^{env}_{t+1}$, which we denote it by .
Fig. \ref{fig:dm} shows the difference learning module. One issue with this model is that, for incorrect predictions during training, adding $\hat{\mathbf{j}}^{\text{diff}}_{t+1}$ to the input frame $\mathbf{j}^{ego}_{t+1}$  causes $\hat{\mathbf{j}}^{env}_{t+1}$ to go out of the range of the input. This is especially problematic for multi-step prediction when this error is accumulated. A similar structure has been suggested in \cite{mohajerin2019multi} for predicting binary OGMs, where a classifier layer is used after the summation to take the result within the input range. Although this can potentially resolve the issue, adding such classifier makes the learning process very slow. This is because the classifier ideally acts as a (sigmoid-shaped) clipper  and the derivative of the clipper for the out of range values, caused by the actual error, is very small.  In our model, we do not use these additional layers. The first term of our reconstruction loss remains the same as before and for the SSIM  part we simply clip the prediction $\hat{\mathbf{j}}^{env}_{t+1}$ and then compute the SSIM: 
\begin{equation}
     \mathcal{L}_t^{\text{rec.}} = \mathcal{D}(\mathbf{j}^{env}_{t+1} , \hat{\mathbf{j}}^{env}_{t+1}) + \lambda \big ( 1 - \text{SSIM} (\mathbf{j}^{env}_{t+1},\textit{clip}(\hat{\mathbf{j}}^{env}_{t+1})) \big ).
\end{equation}
This enables us to backpropagate the incorrect difference predictions through the $\mathcal{D}(\mathbf{j}^{env}_{t+1} , \hat{\mathbf{j}}^{env}_{t+1})$ term and also to make use of the SSIM term.  For future time-steps the clipped output is fed back to the network. 
% %
%  \begin{figure}[!t] 
%     \centering
%     \includegraphics[trim= 5mm 20mm 5mm 10mm, width = 7cm]{prediction_module.pdf}
%     \caption{\small  Prediction module. The top green block is the deterministic encoder that is shared between $p_{\psi}(.)$ and $q_{\phi}(.)$. The bottom part is the stochastic part. Yellow and red block belong to $p_{\psi}(.)$ and $q_{\phi}(.)$, respectively. The switch controls  training vs. inference. }
%     \label{fig:pm}
% \end{figure} 
% %
%%%%%%%%%%%%%%%%%%%%%%%%%%%%%%%%%%%%%%%

\noindent \textbf{Motion Encoding: } To further enhance the predictive power of our model we capture the past motion of other agents in the scene by encoding the difference between consecutive OGMs in the input sequence.  In fact, we built a sequence of $\mathbf{j}^{\text{diff}}_{1:t}$ from the input observations and encode it as a part of the shared code. While $\mathbf{i}_{1:t}$ contain information about motion of other agents relative to the ego-vehicle, $\mathbf{j}^{\text{diff}}_{1:t}$ represent their absolute motion and therefore encoding $\mathbf{j}^{\text{diff}}_{1:t}$ allows reasoning about higher level motion features such as intention of other agents. We model these features by a Gaussian distribution $\mathcal{N}(\mu(\mathbf{j}^{\text{diff}}_{1:t}),\epsilon \mathbf{I})$, where $\mu(\mathbf{j}^{\text{diff}}_{1:t})$ is a neural network and $\epsilon$ is a constant ($\epsilon = 0.5$ in the experiments). Motion encoding is used in both the base model and the DL variant. 

Fig. \ref{fig:pm_h} provides a high level architecture of the model. The implementation details of the prediction module are provided in the supplementary material.

\begin{table*}[!t] 
\vspace{-.4cm}
\scriptsize
\centering
  \begin{tabular}{l||c c c c||C{1.8cm} C{1.8cm} C{1.8cm} C{1.8cm}}
  \hline
   Dataset  $\rightarrow$  &  \multicolumn{4}{c||}{NGSIM I-80} &  \multicolumn{4}{c}{Argoverse} \\
   \hline
   \multirow{2}{*}{Method} &
      \multicolumn{4}{c||}{MSE} &
    TP \hspace{.6cm} TN & TP \hspace{.6cm} TN & TP\hspace{.6cm} TN & TP \hspace{.6cm} TN  \\
         \cline{2-9}
    & $k = 1$ & $k= 5$ 	& $k = 10$ & $k= 20$ &  $k = 1$ & $k= 5$ 	& $k = 10$ & $k= 20$  	\\ 
	\hline
	\hline
     {FM-MPUR}  &  3.4  $\pm$ 0.1 &  4.7  $\pm$ 0.2  & 5.2  $\pm$   0.2 & 7.8  $\pm$ 0.1 &  { 96.24 } \hspace{.2cm}  { 99.68 }&  { 88.44 } \hspace{.2cm}  { 97.12 } &  { 74.62 } \hspace{.2cm}  { 94.21 }&  { 65.60 } \hspace{.2cm}  { 85.31 }\\
    \hline
    
         {RNN-Diff2.1}  &  4.2  $\pm$  0.2  &  14.5  $\pm$  0.3  &  36.6   $\pm$  1.1 &  80.2   $\pm$  2.0&  { 99.21 } \hspace{.25cm}  { \textbf{99.91} }&  { 93.19 } \hspace{.25cm}  { \textbf{99.85} } &  { 87.23 } \hspace{.2cm}  { 99.27 }&  { 80.55 } \hspace{.2cm}  { 93.92 }\\
  
    \hline
    
             {PA}  & \textbf{2.9}  $\pm$ \textbf{0.2} & \textbf{4.1} $\pm$  \textbf{0.1}&  \textbf{4.7}  $\pm$ \textbf{ 0.1} & \textbf{6.2} $\pm$ \textbf{0.2}  &  { 99.18 } \hspace{.2cm}  { 99.89 }&  { 94.12 } \hspace{.2cm}  { 99.76 } &  { 90.14 } \hspace{.2cm}  { 99.48}&  { 83.17 } \hspace{.2cm}  { 96.87 }\\

             {PA-DL}  &  3.3   $\pm$ 0.1 &  4.6 $\pm$  0.1 &  5.1 $\pm$ 0.2  & 6.7 $\pm$ 0.3 &  { \textbf{99.40} } \hspace{.25cm}  \textbf{ 99.91 }&  { \textbf{97.13} } \hspace{.25cm}  { 99.83 } &  { \textbf{92.55} } \hspace{.3cm}  { \textbf{99.71} }&  { \textbf{87.98} } \hspace{.3cm}  { \textbf{98.02} }\\
    \hline
  \end{tabular}
  \vspace{-.3cm}
  \caption{\small Comparison of different models in terms of MSE for NGSIM I-80 and TP/TN for Argoverse. For this table predictions for all methods except RNN-Diff2.1 are generated using the mean value for the latent code.}
  \label{table:tb1}
\end{table*}

%%%%%%%%%%%%%%%%%%%%%%%%%%%%%%%%%%%%%%%%%%%%%%%%%%%%%%%%%%%%%%%%%%%%
\begin{table*}[!t] 
\vspace{-.3cm}
\scriptsize
\centering
  \begin{tabular}{l||C{1.45cm} C{1.45cm} C{1.45cm} C{1.45cm}||C{1.45cm} C{1.45cm} C{1.45cm} C{1.45cm}}
  \hline
   Dataset $\rightarrow$ &  \multicolumn{4}{c||}{NGSIM I-80} &  \multicolumn{4}{c}{Argoverse} \\
   \hline
   Method 
    & $k = 1$ & $k= 5$ 	& $k = 10$ & $k= 20$ &  $k = 1$ & $k= 5$ 	& $k = 10$ & $k= 20$  	\\ 
	\hline
	\hline
     {FM-MPUR}  &  212.6   $\pm$ 5.3  &  207.1 $\pm$ 6.9 &  201.4 $\pm$ 5.1 & 187.3   $\pm$ 7.5 &  624.3 $\pm$ 8.1 &  609.5    $\pm$ 5.6&  538.4  $\pm$ 6.2&  461.7  $\pm$  4.5\\
    \hline
    
       {RNN-Diff2.1}  &  194.2   $\pm$ 7.1&  141.3   $\pm$ 3.6&  71.1  $\pm$ 3.4& 26.6   $\pm$ 1.1 &  656.2   $\pm$ 2.9  &  631.0  $\pm$ 3.0 &  603.1   $\pm$ 5.6 &  545.6  $\pm$  5.8\\
        \hline
    
         {PA}  &  \textbf{236.9}  $\pm$ \textbf{2.7} &  \textbf{232.66}   $\pm$ \textbf{3.9}&  \textbf{225.4}   $\pm$ \textbf{2.4} &  \textbf{211.2}   $\pm$ \textbf{5.2} & 661.1   $\pm$ 6.7 &  657.1  $\pm$ 7.9 &  635.2  $\pm$ 5.3  &  590.7   $\pm$ 4.2\\

         {PA-DL}  &  221.7  $\pm$ 1.7  &  217.5  $\pm$  3.1  &  210.2 $\pm$ 4.8  & 202.9 $\pm$ 5.1   &  \textbf{674.6}   $\pm$ \textbf{2.2}&  \textbf{660.2}  $\pm$ \textbf{2.2}  &  \textbf{642.2}  $\pm$ \textbf{3.9} &  \textbf{603.4}  $\pm$  \textbf{3.7}\\
    \hline
  \end{tabular}
  \vspace{-.3cm}
  \caption{\small Comparison of different models in terms of ALL.}
  \label{table:tb2}
\end{table*}

%%%%%%%%%%%%%%%%%%%%%%%%%%%%%%%%%%%%%%%%%%%%%%%%%%%
\section{Experiments}
In this section, we evaluate the performance of our model, i.e. prediction  by anticipation and its difference learning extension, referred to as PA and PA-DL, respectively.

\noindent\textbf{Baselines:} The proposed  model is an OGM-in OGM-out model and therefore we compare its performance with two similar prediction models, which are, to the best of our knowledge, the state-of-the-art for this unsupervised prediction task: 
\begin{itemize}
\itemsep-.3em
    \item \textbf{Forward Model in Model-Predictive Policy Learning with Uncertainty Regularization  (FM-MPUR)} \cite{henaff2019model}:  FM-MPUR is a CVAE-based model, which aims to directly maximize the log-likelihood $\log p(\mathbf{o}_{t+1:t+k}|\mathbf{o}_{1:t}, \mathbf{a}_{t:t+k-1})$.  Latent code of the FM-MPUR model has an unconditioned prior. Therefore, latent samples are independent of the input frames. This can potentially hurt the prediction accuracy for longer horizons. 
    
    \item \textbf{RNN-based model with Difference Learning component (RNN-Diff)} \cite{mohajerin2019multi}: RNN-Diff is an encoder-decoder structure that uses an RNN in the code space. Encoding and decoding are done using convolutional layers. The main idea in RNN-Diff is removing the ego-actions for a whole sequence of frames as if the ego-vehicle does not move and the scene is observed by a fixed observer for the whole sequence. Therefore they can just focus on predicting the movement of dynamic objects in a fixed scene by learning the difference between consecutive frames. We use the best architecture of their model, named RNN-Diff2.1, for comparison. 
    
\end{itemize}

FM-MPUR takes $20$  frames as its input.
However, we found $10$ input frames to be as rich as $20$  in terms of information about the past, i.e. the Markov property of the sequence is preserved by $10$ frames. Therefore a sum over the conditional log-likelihood can result in a log-likelihood of the whole training set. The RNN-Diff2.1 model also uses $10$ input frames. 

\noindent \textbf{Metrics:} For real-valued OGMs we report \textit{MSE}. 
For binary OGMs we assign class 1 (positive) to occupied pixel and class 0 (negative) to free pixels and report the results in terms of classification scores, i.e. \textit{true positive (TP) and true negative (TN)}. This enable us to distinguish between the accuracy of predicting occupied and free pixels. TP is the more important metric for safe driving as it shows how well a model predicts the obstacles in the environment. 
We also report the results in terms of \textit{average log-likelihood (ALL)}.  In fact, we use kernel density estimation (KDE) by approximating the pdf of the training data using 10K training samples and then evaluating the approximated pdf on the predicted frames of the test sequences with different prediction horizons. We use Gaussian kernel with $\sigma=0.1$. 

%%%%%%%%%%%%%%%%%%%%%%%%%%%%%%%%%%%%%%%%%%%%%%%%%%%%%%%%%%%%%%%%%%%%%%%%%%%%%%%%%%%%%
\subsection{Prediction under different driving situations}
\label{sec:regularaction}
We ran our experiments on two complimentary datasets, one high-speed highway traffic and the other low-speed dense urban traffic. These two datasets cover a large set of real-life driving scenarios.  

\textbf{NGSIM I-80 dataset:}
The  Next Generation Simulation program’s Interstate 80 (NGSIM I-80) \cite{ngsim} dataset consists of 3 batches of 15-minute of recordings from traffic cameras mounted over a stretch of a highway in the US. Driving behaviours are complex with complicated interactions between vehicles moving at high-speed. This makes the future state difficult to predict. We follow the same preprocessing proposed in \cite{henaff2019model} to make the datasets. The images are real-valued RGB with size $117 \times 24$. The ego-vehicle is in the center of the blue channel. Other (social) vehicles are in the green channel, which can also be interpreted as the OGM. The red channel has the map information, e.g. lanes.

\begin{figure}[!t]
    \centering
    \includegraphics[width = 8cm]{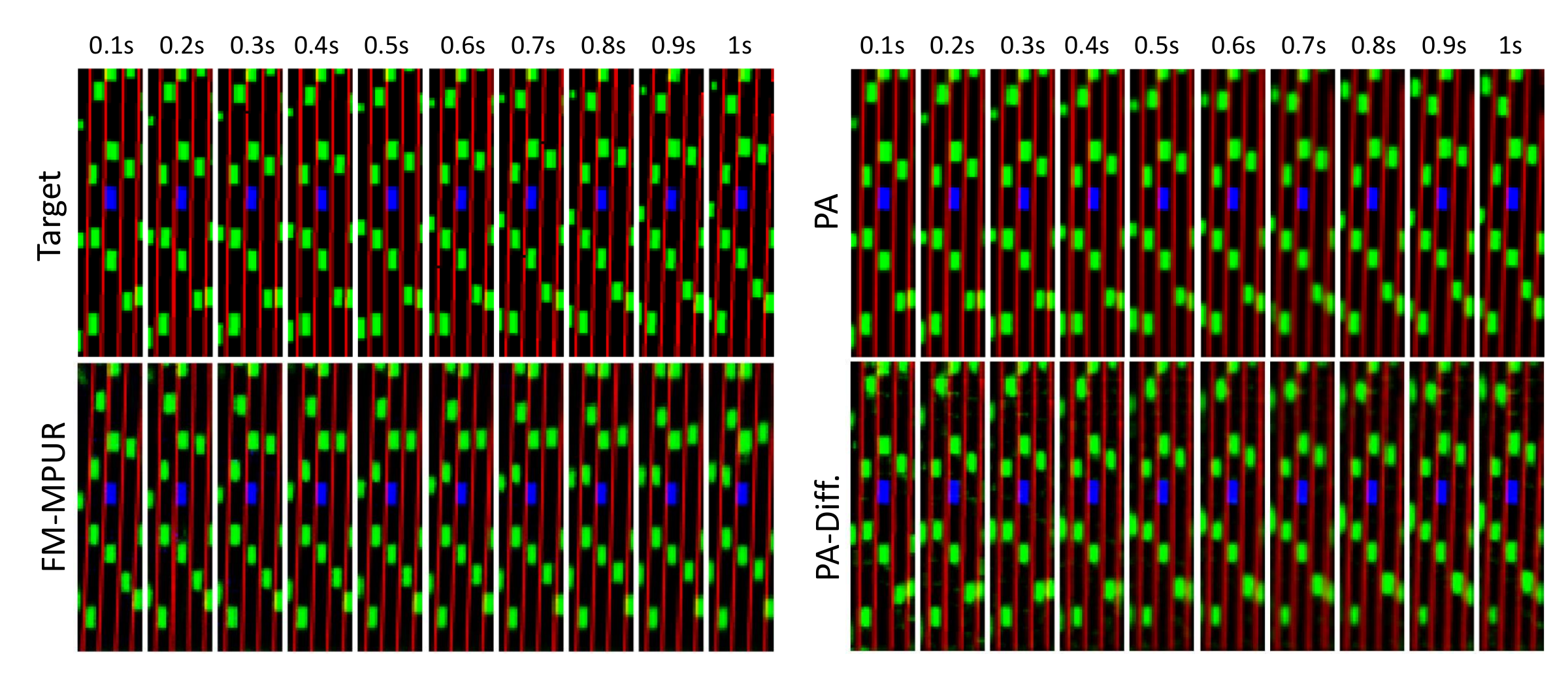}
        \vspace{-.5cm}
    \caption{\small Predictions of 10 frames (1 sec) by different models.
    }
    \vspace{-.6cm}
    \label{fig:ngsim_1}
\end{figure}

We train each model using two batches of the 15-minute recording and test it on the third batch and repeat this process three times to cover all combinations.  For both PA and PA-DL, we put $\lambda = 0.05$ in the reconstruction cost. Results are shown in Tables \ref{table:tb1} and \ref{table:tb2}. Since the speed of the ego-vehicle is high in this dataset, removing the ego-motion for the whole sequence to train the RNN-Diff2.1, significantly reduces the size of meaningful pixels in the input and target frames and make them practically unusable for training a multi-step model. Therefore the reported results are from a model trained for single-step prediction. This is why its performance dramatically drops for larger values of $k$. As we can see, PA and PA-DL outperform FM-MPUR. Due to high-speed driving of agents in this dataset, their positions have dramatic changes from one frame to the next one, in many cases. Consequently, PA-DL performs worse than the base model.  
Fig. \ref{fig:ngsim_1} shows a sequence of predictions for different models to allow us to see PA, PA-DL, and FM-MPUR performances qualitatively.% for $k=10$. 
\begin{figure*}[!t]
\vspace{-.3cm}
    \centering
\includegraphics[trim = 0mm 0mm 0mm 0mm, height=4cm]{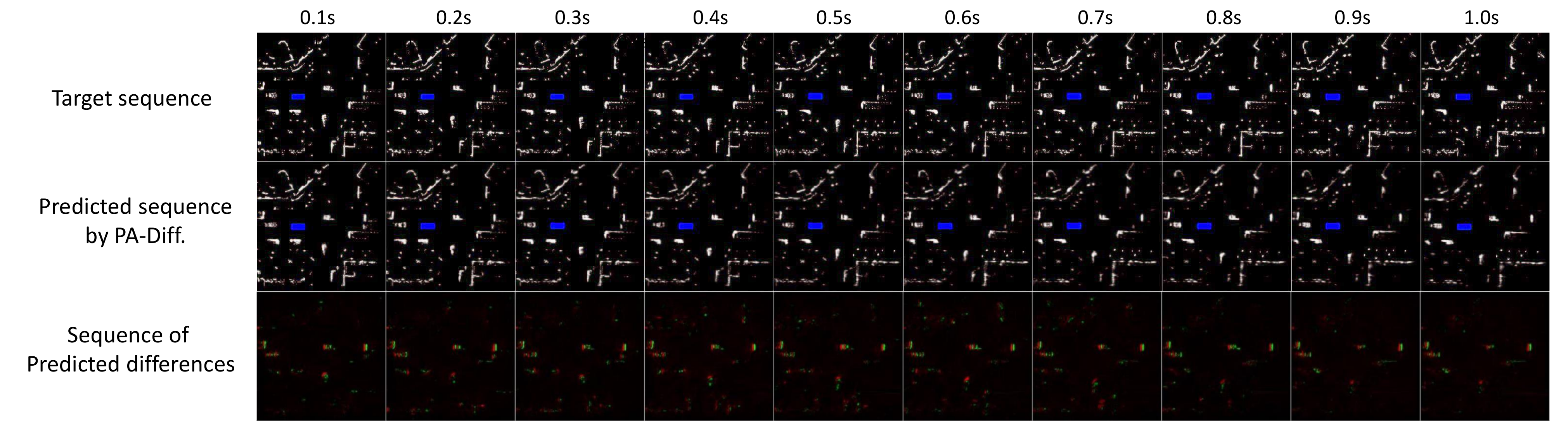}
	  \vspace{-.4cm}
    \caption{\small OGM prediction using PA-DL model. The top row is the target sequence and the middle row shows the predictions. The bottom row shows the difference between two consecutive frames, learned by the model, where the red areas are erased from the frame and the green areas are added to build the next frame. 
    }
    \vspace{-.7cm}
    \label{fig:argo}
\end{figure*}

%
%%%%%%%%%%%%%%%%%%%%%%%%%%%%%%%%%%%%%%%%%%%%%%%%%%%%%%%%%%%%%%%%%%%%%%%%%%%%%%%%%%%%%%%%%%%%%%%%%
\textbf{Argoverse dataset:}
For the urban area driving, the OGM sequences are obtained from the Argoverse raw dataset \cite{chang2019argoverse}. 
The dataset contains many different actions and maneuvers, e.g. stops and turns, in slow pace. The LiDAR point-clouds, collected at 10Hz, are converted to BEV $256 \times  256$ binary OGMs using ground removal proposed in \cite{miller2006mixture}. We compare the performance of the models in terms of TP/TN and ALL in Tables \ref{table:tb1} and \ref{table:tb2}, respectively. Argoverse dataset has more complicated OGM structures and, unlike NGSIM I-80, the ego-motion is usually small and consecutive frames have slight differences. Therefore, PA-DL outperforms PA, and both PA and PA-DL outperform FM-MPUR significantly. 

 PA-DL and RNN-Diff2.1 perform closely for short-horizon predictions. Again, since in RNN-Diff2.1 the environment is observed from a fixed point, when $k$ is large the dynamic objects eventually leave the scene and the predictions deviate from the actual ground truth. In PA-DL, the difference learning is done step-by-step. As we can see, the performance gap between RNN-Diff2.1 and PA-DL enlarges as $k$ grows. Fig. \ref{fig:argo} shows a sample sequence of the Argoverse dataset as well as the outputs of the PA-DL model. 
 
The results of Tables \ref{table:tb1} and \ref{table:tb2} show that the proposed models outperform the two baselines with a significant margin. 
%Although the performance gap for a single driving situation might be marginal, our models can predict well on both datasets. In contrast, 
Moreover, each of the baselines fails in one of the two datasets, while PA and PA-DL perform the prediction task successfully for both datsets, i.e. both driving situations.

\subsection{Prediction for rare actions} 
 %%%%%%%%%%%%%%%%%%%%%%%%%%%%%%%%%%%%%%%%%%%%%%%%%%%%%%%%%
 %likelihood table
\begin{table*}[!t] 
\scriptsize
\centering
  \begin{tabular}{l||C{2.5cm} C{2.5cm} C{2.5cm} C{2.5cm}}
  \hline
 
      \multirow{2}{*}{Method} &
      \multicolumn{4}{c}{ALL}  \\
         \cline{2-5}
 & $k = 1$ & $k= 5$ 	& $k = 10$ & $k= 20$  		\\ 
	\hline
	\hline
FM-MPUR  & 197.3 $\pm$  6.2& 175.7 $\pm$  5.1& 112.1 $\pm$  7.7& 76.4 $\pm$ 4.1\\
    \hline
PA & $\textbf{226.1} \pm \textbf{2.3}$ & $\textbf{218.6} \pm \textbf{4.1}$& $\textbf{202.2} \pm \textbf{3.8}$&$\textbf{180.4} \pm \textbf{6.6}$\\

PA-DL &  {213.4}  $\pm$   {6.8} &  {200.4}  $\pm$   {4.9}&  {192.9}  $\pm$   {3.5}& { 174.8}  $\pm$   {7.9}\\
    \hline
  \end{tabular}
  \vspace{-.35cm}
  \caption{\small comparison of predictions of PA, PA-DL, and FM-MPUR using rare actions.}
  \label{table:tb3}
  \vspace{-.1cm}
\end{table*}

 %%%%%%%%%%%%%%%%%%%%%%%%%%%%%%%%%%%%%%
In this section we study the performance of the PA and PA-DL algorithms in the presence of rare actions and investigate the effectiveness of employing prior knowledge in providing robustness against the actions that are rare in the training data. Specifically we consider the NGSIM I-80 dataset.

\begin{figure}[!t]
    \centering
    \includegraphics[trim = 0mm  0mm  0mm  0mm, width = 7.3cm]{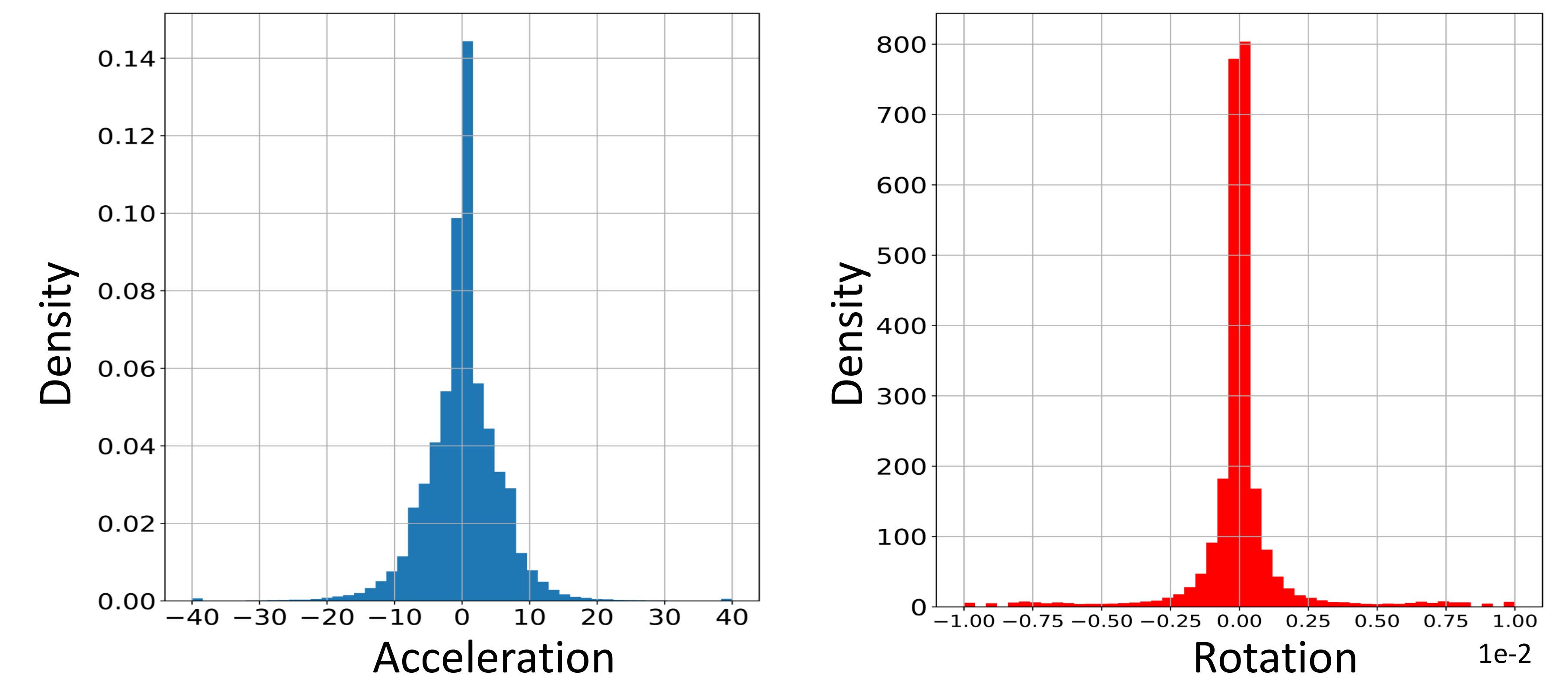}
    \vspace{-.3cm}
    \caption{\small Distribution of actions.
    }
    \vspace{-.4cm}
    \label{fig:actions}
\end{figure}

 We use the trained models with each of the batches of 15-minute recordings and apply actions that are rarely seen in the training set but are still in the maneuverability range of vehicles. We use the distributions shown in Fig. \ref{fig:actions} to sample these actions and apply them to randomly selected sequences of the test set and predict for different prediction horizons. For comparison, we use the FM-MPUR algorithm. Since there is no ground truth, we only report the ALL results.
% For evaluation we use two different metrics. First, we count the number of visually corrupted images and call them invalid predictions. 
% To remove the human bias, we also compute ALL using KDE with the same procedure described in section \ref{sec:regularaction}.
%%%%%%%%%%%%%%%%%%%%%%%%%%%%%%%%%%%%%%%%%%%%%%%%%%%%%%%%%%%%%%%%%%%%%%%%%%%%%%%%%%%%%%
\begin{table*}[!t] 
\vspace{-.2cm}
\scriptsize
\centering
  \begin{tabular}{l||c c c c||C{1.8cm} C{1.8cm} C{1.8cm} C{1.8cm}}
  \hline
   Dataset  $\rightarrow$  &  \multicolumn{4}{c||}{NGSIM I-80} &  \multicolumn{4}{c}{Argoverse} \\
   \hline
   \multirow{2}{*}{Method} &
      \multicolumn{4}{c||}{MSE} &
    TP \hspace{.6cm} TN & TP \hspace{.6cm} TN & TP\hspace{.6cm} TN & TP \hspace{.6cm} TN  \\
         \cline{2-9}
    & $k = 1$ & $k= 5$ 	& $k = 10$ & $k= 20$ &  $k = 1$ & $k= 5$ 	& $k = 10$ & $k= 20$  	\\ 
	\hline
	\hline
     {PA (no RBM)}  &  3.2  $\pm$ 0.2 &  4.5  $\pm$ 0.2  & 5.0  $\pm$ 0.1 &7.2  $\pm$ 0.4  &  {97.03 } \hspace{.25cm}  { 99.74 }&  { 90.15 } \hspace{.25cm}  { 97.98 } &  { 80.12 } \hspace{.25cm}  {95.23 }&  {72.12} \hspace{.25cm}  { 88.63 }\\

         {PA (no BCDE)}  &  3.1  $\pm$  0.3  &  4.2  $\pm$  0.4  &  4.7   $\pm$  0.2 &  6.9   $\pm$  0.3&  { 99.01 } \hspace{.25cm}  {  {99.84} }&  { 93.20 } \hspace{.25cm}  {  {99.64} } &  { 88.18 } \hspace{.25cm}  { 99.22 }&  { 80.14 } \hspace{.25cm}  { 92.20}\\

             {PA (no ME)}  & 3.1  $\pm$  {0.2} &  {4.3} $\pm$   {0.4}&   {4.8}  $\pm$  { 0.3} &  {6.7} $\pm$  {0.2}  &  { 98.65 } \hspace{.25cm}  { 99.75 }&  { 92.51 } \hspace{.25cm}  { 98.50 } &  {86.19 } \hspace{.25cm}  { 98.98 }&  { 81.35 } \hspace{.25cm}  { 92.71 }\\

          {PA-DL (no RBM)}  &  4.1   $\pm$ 0.4 &  6.9 $\pm$  0.3 &  7.3 $\pm$ 0.1  &10.8 $\pm$ 0.2 &  {  {95.51} } \hspace{.25cm}  { 99.83 }&  {  {91.94} } \hspace{.25cm}  { 97.72 } &  {  {81.25} } \hspace{.25cm}  {  {95.74} }&  {  {70.55} } \hspace{.25cm}  {  {86.30} }\\

             {PA-DL (no BCDE)}  & 3.4   $\pm$ 0.2 &  4.7  $\pm$  0.1 &  5.6 $\pm$ 0.2  & 8.4 $\pm$ 0.2 &  {  {99.20} } \hspace{.25cm}  { 99.88 }&  {  {96.95} } \hspace{.25cm}  {99.70 } &  {  {90.62} } \hspace{.25cm}  {  {99.51} }&  {  {83.24} } \hspace{.25cm}  {  {91.95} }\\
             
          {PA-DL (no ME)}  &  3.6    $\pm$  0.2 &  4.7  $\pm$  0.1 &  5.4 $\pm$ 0.2  & 8.1 $\pm$ 0.3 &  {  {99.05} } \hspace{.25cm}  { 99.80 }&  {  {96.54} } \hspace{.25cm}  { 99.71 } &  {  {91.14} } \hspace{.25cm}  {  {99.59} }&  {  {84.64} } \hspace{.25cm}  {  {94.34} }\\
    \hline
  \end{tabular}
  \vspace{-.3cm}
  \caption{\small Results of ablative study on the contributing factors to the performance of our models. no RBM: a model without rule-based modules. no BCDE: a CVAE-based model with unconditioned prior for the latent code. no ME: a module without motion encoding. }
  \label{table:ablation-msetptn}
    \vspace{-.4cm}
\end{table*}

\begin{figure}[!t]
    \centering
	\includegraphics[trim = 20mm 0mm 20mm 0mm, width=7cm]{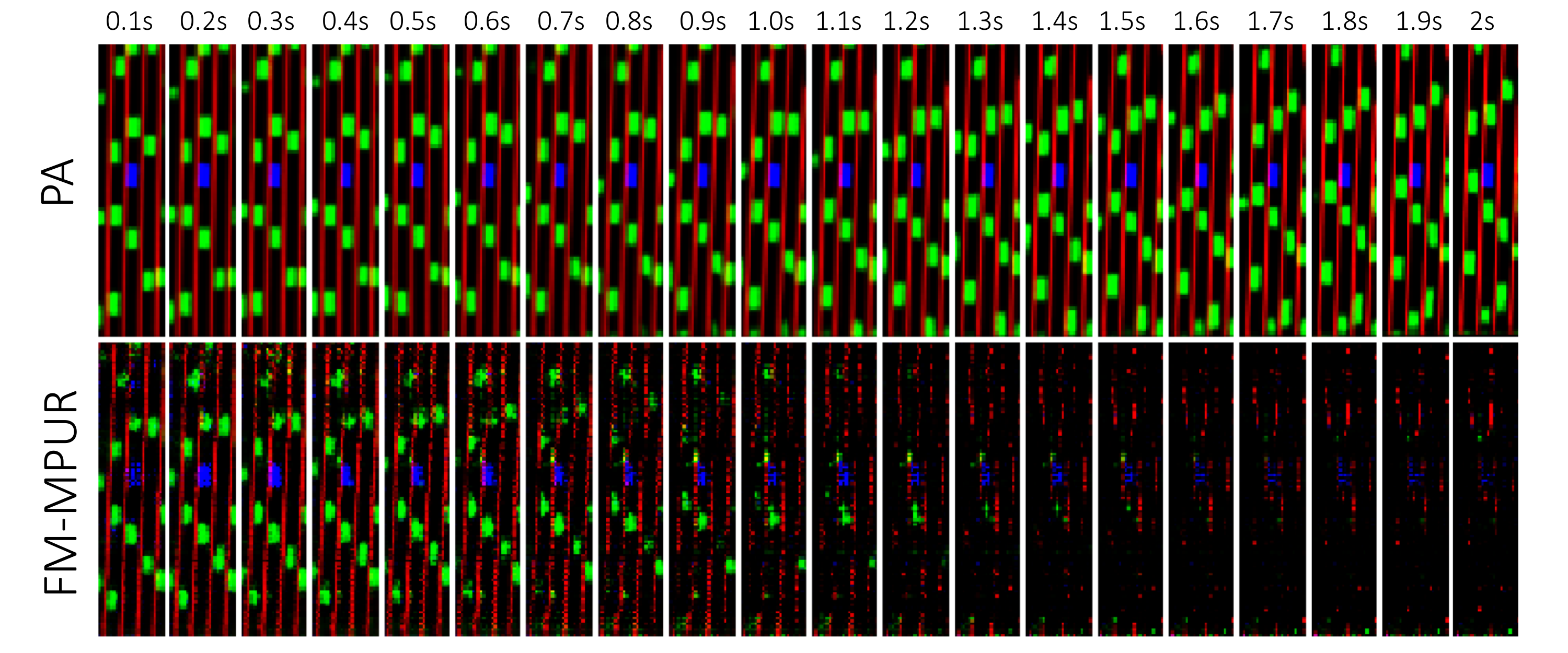}
	\vspace{-.4cm}
    \caption{\small Effect of constantly applying rare actions on the prediction of PA and FM-MPUR models. 
    }
    \vspace{-.55cm}
    \label{fig:ngsim_2}
\end{figure}

%%%%%%%%%%%%%%%%%%%%%%%%%%%%%%%%%%%%%%%%%%%%%%%%%%%%%%%%%%%%%%%%%%%%%%%%%%%%%%%%%%%%%%
Table \ref{table:tb3} summarizes the evaluation results. It shows that our models significantly outperform FM-MPUR for this task, especially for longer prediction horizons. This suggests that learning environment-features based on the anticipated ego-features makes the prediction task easier to learn for our model, which supports our initial intuition. In fact, by applying the extreme actions, the OGMs change dramatically from one time step to the next one. However, we can compensate this change by applying the anticipated modifications to the target OGM. Fig.~\ref{fig:ngsim_2} shows the result of applying rare actions to the same input sequence in Fig. \ref{fig:ngsim_1}. We apply $\mathbf{a} = [-25 , 0]$ for $20$ consecutive steps, which can be identified as a very low-probable action sequence according to the distributions in Fig. \ref{fig:actions}. This is equivalent to a hard brake in the middle of the road. As we can see our model can predict almost perfectly, while the FM-MPUR model fails after a few predictions. Compared to the predictions that correspond to the original actions in Fig. \ref{fig:ngsim_1}, location of the nearby social vehicles show that the model has learned the dynamics of the traffic: as the ego-vehicle brakes hard, other vehicles continue to move normally except for the one behind it, which is forced to slow down significantly. 
%
%%%%%%%%%%%%%%%%%%%%%%%%%%%%%%%%%%%
\subsection{Ablation study}
There are three main factors contributing to the better performance of our model compared to the baselines: 1) Employing prior knowledge using rule-based modules to set up the \textit{anticipation-interaction} training. 2) Employing the bottleneck model that conditions the prior of the latent code on input. 3) Encoding \textit{absolute} motion of other agents. We conduct an ablative study on each of these factors for both PA and PA-DL models and provide the results in Table \ref{table:ablation-msetptn} based on MSE and TP/TN . Comparing the results of this table with Table \ref{table:tb1}, we can see that while all factors are contributing, the rule-based modules contribute more. Since the difference learning is a byproduct of our main idea of, removing the rule-based modules degrades the performance of PA-DL significantly.  Also motion encoding plays a slightly more important role than the BCDE (conditioned prior) model.  The effect of using conditioned prior becomes more apparent for larger values of $k$. 

ALL results for both regular and low-probable action settings are provided in the supplementary materials, which show similar behaviors. We also present an ablation study on the effect of the SSIM term on the reconstruction loss in the supplementary materials.

% \noindent \textbf{SSIM term: } The SSIM term helps to generate more clear OGMs. Using the validation set the optimal weights for this term, denoted by $\lambda^*$, are $\lambda^* = 0.05$ for the NGSIM I-80 dataset and $\lambda^* = 0.1$ for the Argoverse dataset. This holds for both PA and PA-DL models. 
% A comparison between  $\lambda =0$, i.e. not using the SSIM term, and $\lambda^*$ is presented in Tables \ref{table:ssim1} and  \ref{table:ssim2}.
% As we can see in both cases, adding the the SSIM term is effective, especially for larger values of $k$. This is due to the fact that the error of the prediction is accumulative.  

%%%%%%%%%%%%%%%%%%%%%%%%%%%%%%%%%%%
\vspace{-.1cm}
\section{Conclusion}
\vspace{-.1cm}
We proposed that an observed interaction sequence can be explained by an underlying generative process wherein some agents act partly in response to the anticipated action of other agents. Based on this view, we factorized the interaction sequence into anticipated action and anticipated partial reaction, thereby setting up an action-conditional distribution. We designed a bottleneck conditional density estimation model to learn the distribution. In comparison to the baselines, our model achieves a higher capacity for prediction: it reaches higher accuracy, it handles rare actions much better, it is able to perform well under different driving situations, including high-speed highway driving and complicated urban navigation. While our experiments are limited to vehicle-vehicle interaction, insofar as the understanding generative process is pertinent, our method may also generalize well to other tasks, such as prediction of pedestrian-vehicle interaction, or to other multi-agent domains. Finally, because our model is action conditional, it can serve as a world model for many downstream tasks.

{\small
\bibliographystyle{ieee_fullname}
\bibliography{main}
}
\newpage
\appendix
\onecolumn
\begin{center}
    \Large \textbf{Prediction by Anticipation: An Action-Conditional Prediction Method based on Interaction Learning \\Supplementary Material}
\end{center}{}
%\centering{\section*{Supplementary Materials}}
\section{Terms in the loss function}
We re-write the loss function in Eq. \ref{eq: objective} here: 

\begin{align}
    \nonumber
     \mathcal{L}_t &=  
     \underbrace{\mathcal{D}(\mathbf{j}^{env}_{t+1} , \hat{\mathbf{j}}^{env}_{t+1}) \!+ \!\lambda \big( 1- \text{SSIM} (\mathbf{j}^{env}_{t+1},\hat{\mathbf{j}}^{env}_{t+1}) \big)}_{\mathcal{L}_t^{\text{rec.}}}  + \underbrace{\text{KL} \big (q_{\phi}(\mathbf{z}_t|\mathbf{o}_{1:t}, \mathbf{j}^{ego}_{t+1}, \mathbf{j}^{env}_{t+1} ) || p_{\psi}(\mathbf{z}_t|\mathbf{o}_{1:t}, \mathbf{j}^{ego}_{t+1} ) \big )}_{\mathcal{L}_t^{\text{KL}}}. 
    \label{eq: objective}
\end{align}

\subsection{The ELBO term}  
Minimizing the first and last terms is equivalent to maximizing the ELBO in Eq. \ref{eq: lowerbound}.  Considering $q_{\phi}(\mathbf{z}_t|\mathbf{o}_{1:t}, \mathbf{j}^{ego}_{t+1}, \mathbf{j}^{env}_{t+1} )= \mathcal{N}(\mu_{\phi}, \Sigma_{\phi})$ and $p_{\psi}(\mathbf{z}_t|\mathbf{o}_{1:t}, \mathbf{j}^{ego}_{t+1} ) = \mathcal{N}(\mu_{\psi}, \Sigma_{\psi})$ as Gaussian distributions, the KL term is simply differentiable with the following terms:
\begin{align*}
\text{KL} \big (q_{\phi}(\mathbf{z}_t|\mathbf{o}_{1:t}, \mathbf{j}^{ego}_{t+1}, \mathbf{j}^{env}_{t+1} ) ) || p_{\psi}(\mathbf{z}_t|\mathbf{o}_{1:t}, \mathbf{j}^{ego}_{t+1} ) \big ) = \frac{1}{2} \Big(\text{Tr}\big({\mathbf{\Sigma}_{\psi}}^{-1}\mathbf{\Sigma}_{\phi}\big) + \big(\mu_{\psi} - \mu_{\phi}\big)^{\top} {\mathbf{\Sigma}_{\psi}}^{-1} \big(\mu_{\psi}- \mu_{\phi}\big)+ \log (\frac{|\mathbf{\Sigma}_{\psi}|}{|\mathbf{\Sigma}_{\phi}|}) -d\Big), 
\end{align*}
where $d$ is the dimensionality of the stochastic latent code. 

\subsection{SSIM term and ablation}
The SSIM term helps to generate more clear images. Using the validation set the optimal weights for this term, denoted by $\lambda^*$, are $\lambda^* = 0.05$ for the NGSIM I-80 dataset and $\lambda^* = 0.1$ for the Argoverse dataset. This holds for both PA and PA-DL models. 

The SSIM loss is an auxiliary term in our loss function that does not necessary appear in ELBO. Therefore, we present an ablation study on this term for each of the datasets and report the result in terms of MSE and TP/TN for the case we don't use SSIM in the loss function. A comparison between  $\lambda =0$, i.e. not using the SSIM term, and $\lambda^*$ is presented in Table \ref{table:ssim1} .

%%%%%%%%%%%%%%%%%%%%%%%%%%%%%%%%%%%%%%%%%%%%%%%%%%%%%%%
\begin{table*}[!h] 
\scriptsize
\centering
  \begin{tabular}{l||c c c c||C{1.8cm} C{1.8cm} C{1.8cm} C{1.8cm}}
  \hline
   Dataset  $\rightarrow$  &  \multicolumn{4}{c||}{NGSIM I-80} &  \multicolumn{4}{c}{Argoverse} \\
   \hline
   \multirow{2}{*}{Method} &
      \multicolumn{4}{c||}{MSE} &
    TP \hspace{.6cm} TN & TP \hspace{.6cm} TN & TP\hspace{.6cm} TN & TP \hspace{.6cm} TN  \\
         \cline{2-9}
    & $k = 1$ & $k= 5$ 	& $k = 10$ & $k= 20$ &  $k = 1$ & $k= 5$ 	& $k = 10$ & $k= 20$  	\\ 
	\hline
	\hline
    
             {PA ($\lambda^*$)}  &  {2.9}  $\pm$   {0.2} &  {4.1} $\pm$    {0.3}&   {4.7}  $\pm$   { 0.3} &  {6.2} $\pm$   {0.6}  &  { 99.18 } \hspace{.25cm}  { 99.89 }&  { 94.12 } \hspace{.25cm}  { 99.76 } &  { 90.14 } \hspace{.25cm}  { 99.48}&  { 83.17 } \hspace{.25cm}  { 96.87 }\\

              {PA ($\lambda$ = 0)}  &  {2.9}  $\pm$   {0.3} &  {4.0} $\pm$    {0.3}&   {4.9}  $\pm$   { 0.2} &  {6.5} $\pm$   {0.4}  &  { 99.09 } \hspace{.25cm}  { 99.88 }&  { 94.03 } \hspace{.25cm}  { 99.64 } &  { 87.56 } \hspace{.25cm}  { 98.19}&  { 82.25 } \hspace{.25cm}  { 94.03 }\\

             {PA-DL ($\lambda^*$)}  &  3.3   $\pm$  0.2 &  4.6 $\pm$   0.3 &  5.1 $\pm$  0.4  & 6.7 $\pm$  0.6 &  {  {99.40} } \hspace{.25cm}   { 99.91 }&  {  {97.13} } \hspace{.25cm}  { 99.83 } &  {  {92.55} } \hspace{.25cm}  {  {99.71} }&  {  {87.98} } \hspace{.25cm}  {  {98.02} }\\
             
          {PA-DL ($\lambda$ = 0)}  &  3.4   $\pm$  0.2 &  4.6 $\pm$   0.2 &  5.2 $\pm$  0.4  & 6.9 $\pm$  0.4 &  {  {99.04} } \hspace{.25cm}   { 99.84 }&  {  {96.35} } \hspace{.25cm}  { 99.51 } &  {  {90.99} } \hspace{.25cm}  {  {99.47} }&  {  {85.15} } \hspace{.25cm}  {  {96.13} }\\
    \hline
  \end{tabular}
  \caption{\small Effect of the SSIM term in terms of MSE for NGSIM I-80 and TP/TN for Argoverse.}
  \label{table:ssim1}
\end{table*}

%%%%%%%%%%%%%%%%%%%%%%%%%%%%%%%%%%%%%%%%%%%%%%%%%%%%%%%
%%%%%%%%%%%%%%%%%%%%%%%%%%%%%%%%%%%%%%%%%%%%%%%%%%%%%%%
%\input{ssim-all}
%%%%%%%%%%%%%%%%%%%%%%%%%%%%%%%%%%%%%%%%%%%%%%%%%%%%%%%
As we can see in both cases, adding the the SSIM term is effective, especially for larger values of $k$. This is due to the fact that the error of the prediction is accumulative.

%%%%%%%%%%%%%%%%%%%%%%%%%%%%%%%%%%%%%%%%%%%%%%%%%%%%%%%%%%%%%%%%%%%%%%%%%%%%%%
\section{Ablation study for the ALL measure}
We provided the results of ablative study of different contributing factors on MSE and TP/TN in the main body of the paper. Here we provide the results for ALL, which show a similar trend as MSE and TP/TN in Table \ref{table:ablation-msetptn}.
\begin{table*}[!h] 
\vspace{-.3cm}
\scriptsize
\centering
  \begin{tabular}{l||c c c c||c c c c}
  \hline
   Dataset $\rightarrow$ &  \multicolumn{4}{c||}{NGSIM I-80} &  \multicolumn{4}{c}{Argoverse} \\
   \hline
   Method 
    & $k = 1$ & $k= 5$ 	& $k = 10$ & $k= 20$ &  $k = 1$ & $k= 5$ 	& $k = 10$ & $k= 20$  	\\ 
	\hline
	\hline
     {PA (no RBM)}  &  219.4  $\pm$   4.8  &  214.5  $\pm$   5.4  &  207.0  $\pm$   6.6  & 196.7  $\pm$   7.3   &   {632.7}   $\pm$   {9.2}&   {614.5}  $\pm$  4.7  & 589.9  $\pm$   7.2 &   535.9 $\pm$   3.9\\
     
      {PA (no BCDE)}  &  232.5  $\pm$   5.5  &  228.4  $\pm$   4.9  &  218.9  $\pm$   6.2  & 200.1 $\pm$  6.0   &   {657.9}   $\pm$ 4.5&   {650.4}  $\pm$   {5.6}  &   {619.8}  $\pm$   {6.9} &   {573.6}  $\pm$    {7.0}\\

       {PA (no ME)}  &  230.4  $\pm$   4.5  &  224.3  $\pm$   6.9  &  217.6$\pm$  8.0 & 203.9 $\pm$  5.2   &   {651.2}   $\pm$   {3.4}&   {636.2}  $\pm$   {7.2}  &   {608.3}  $\pm$   {3.2} &   {555.1}  $\pm$    {6.7}\\

         {PA-DL (no RBM)}  &  210.5  $\pm$  3.6  &  205.8  $\pm$   5.8  &  196.3 $\pm$  6.8  & 180.5 $\pm$  7.1   &   {645.6}   $\pm$   {6.2}&   {623.3}  $\pm$   5.2  &  601.4  $\pm$   6.5 &   557.2 $\pm$    4.8\\
         
          {PA-DL (no BCDE)}  &  218.5  $\pm$ 3.1  &  214.2  $\pm$  5.1  &  204.6 $\pm$  6.6  & 193.4 $\pm$  6.4   &   {668.2}   $\pm$   6.1&   {653.2}  $\pm$   {4.4}  &   {633.1}  $\pm$   {7.4} &   {585.4}  $\pm$    {7.8}\\

       {PA-DL (no ME)}  &  218.2  $\pm$  5.0 &  211.5  $\pm$   7.2  &  205.8 $\pm$  4.2  & 195.3 $\pm$  7.5   &   {661.2}   $\pm$   {5.5}&   {641.8}  $\pm$   {7.6}  &   {620.7}  $\pm$   {5.4} &   {577.1}  $\pm$    {5.0}\\
    \hline
  \end{tabular}
  \caption{\small Ablative study in terms of ALL for regular actions.}
  \label{table:tb2}
\end{table*}

\begin{table*}[!h] 
\scriptsize
\centering
  \begin{tabular}{l||C{2.5cm} C{2.5cm} C{2.5cm} C{2.5cm}}
  \hline
 
      \multirow{2}{*}{Method} &
      \multicolumn{4}{c}{ALL}  \\
         \cline{2-5}
 & $k = 1$ & $k= 5$ 	& $k = 10$ & $k= 20$  		\\ 
	\hline
	\hline
PA (no RBM) &  {203.5} $\pm$  {4.5} &  {182.9} $\pm$  {6.3}&  {134.7} $\pm$  {3.4}& { 99.7} $\pm$  {6.2}\\

PA (no BCDE)  &  {221.2} $\pm$  {6.2} &  {211.6} $\pm$  {7.5}&  {190.9} $\pm$  {5.0}& { 161.2} $\pm$  {7.3}\\

PA (no ME) &  {219.7} $\pm$  {6.2} &  {205.2} $\pm$  {6.8}&  {188.2} $\pm$  {7.2}& {163.8} $\pm$  {9.2}\\

PA-DL (no RBM) &  {198.6} $\pm$  {5.4} &  {172.4} $\pm$  {6.8}&  {125.5} $\pm$  {6.2}& {82.3} $\pm$  {4.4}\\

PA-DL (no BCDE)  &  {208.9} $\pm$  {3.9} &  {194.6} $\pm$  {5.6}&  {178.5} $\pm$  {4.9}& { 142.5} $\pm$  {6.8}\\

PA-DL (no ME) &  {209.5} $\pm$  {4.3} &  {191.5} $\pm$  {5.8}&  {180.6} $\pm$  {5.0}& { 149.8} $\pm$  {6.9}\\
    \hline
  \end{tabular}
  \caption{\small Ablative study in terms of ALL for low-probable actions in NGSIM I-80 dataset.}
  \label{table:tb3}
\end{table*}

 It is worth mentioning that the effect of using a conditioned prior for the latent (BCDE model) is more significant for low-probable actions, as the model with conditioned code has access to its previous predictions when estimating the parameter of the latent distribution. 
%%%%%%%%%%%%%%%%%%%%%%%%%%%%%%%%%%%%%%%%%%%%%%%%%%%%%%%%%%%%%%%%%%%%%%%%%%%%
\section{Implementation details of the prediction module}
\begin{figure}[!h]
    \centering
    \includegraphics[width = 16.5cm]{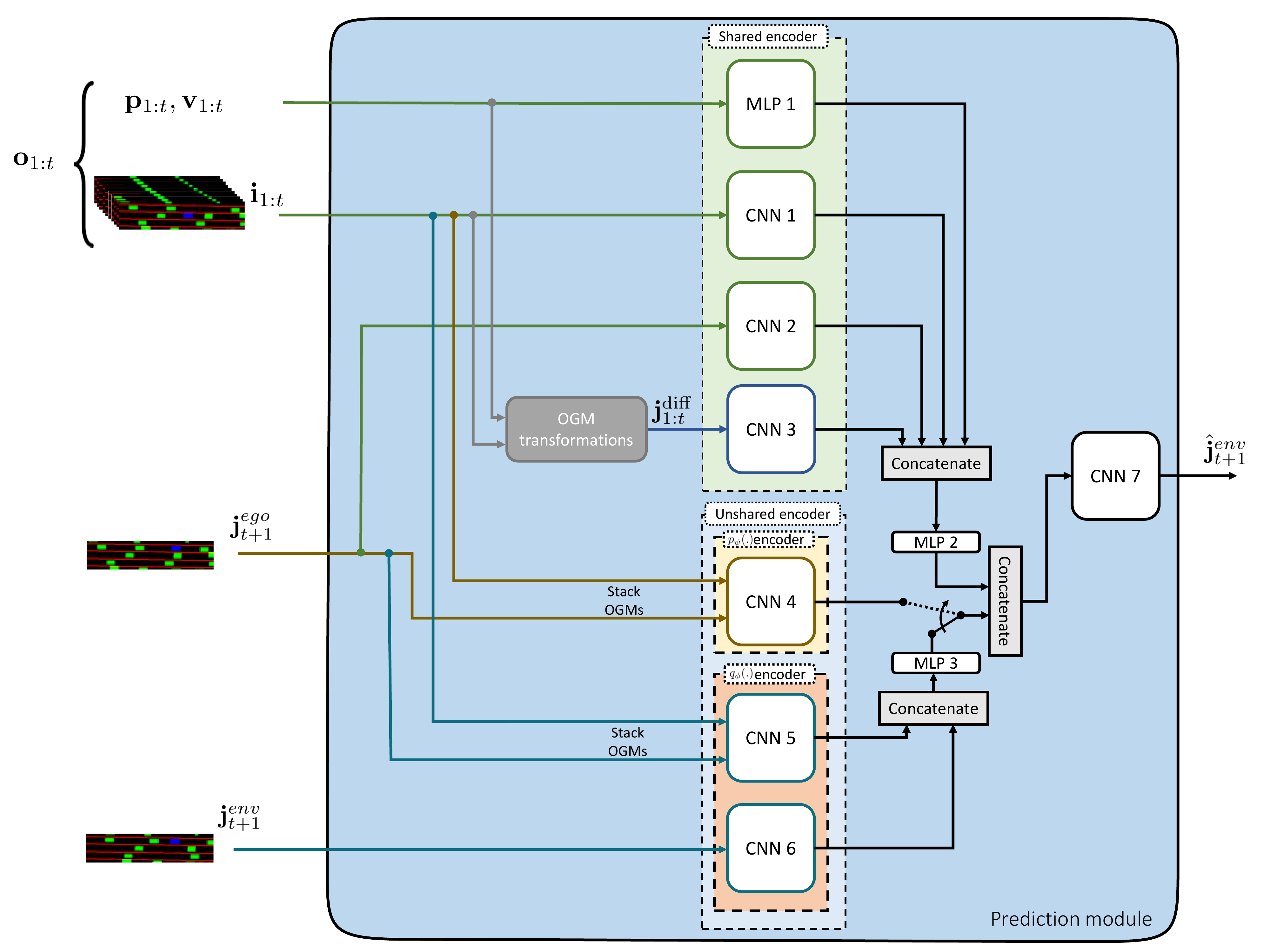}
    \caption{Detailed structure of the prediction module. }
    \label{fig:pm_details}
\end{figure}

Figure \ref{fig:pm_details} shows a detailed implementation of the prediction module. The structure of the networks based on this figure are explained below. Convolutional neural networks, denoted by CNN, have convolutional layers as their core but can also include one or two fully-connected layers. MLP networks only contain fully-connected layers. 

\subsection{NGSIM I-80 dataset}
For the NGSIM I-80 dataset, we use the actions extracted by the authors in \cite{henaff2019model}. As stated in the experiment section, the images are $117 \times 24 \times 3$. For the IOTs we zero pad the images before feeding them to the modules with the padding size of $20$ in each dimension and remove the padding from the output image after processing. 

%%%%%%%%%%%%%%%%%%%%%%%%%%%%%%%%%%%%%%%%%%%%%%%%%%%%%%%%%

\footnotesize{

\begin{longtable}{l||l| L{10.4cm}}
\Xhline{2\arrayrulewidth}
	Network	& \multicolumn{2}{l}{Detail of structure} 	\\ 
\Xhline{2\arrayrulewidth}
%%%%%%%%%%%%%%%%%%%%%%%%%%%%%%%%%%%%%%%%%%%%%%%%%%%%%%%%%%%%
    \multirow{2}{*}{MLP 1} & \multirow{2}{*}{2 layers }& 1. $256$ units- leaky ReLU activation   \\
      
    && 2. $25$ units- ReLU activation\\
   \hline
%%%%%%%%%%%%%%%%%%%%%%%%%%%%%%%%%%%%%%%%%%%%%%%%%%%%%%%%%%%%
    \multirow{4}{*}{CNN 1} & \multirow{4}{*}{4 layers }&1. $64$ kernels of size $4 \times 4$, stride size = $2$, leaky ReLU activation\\
      
    &&2. $128$ kernels of size $4 \times 4$, stride size = $2$, leaky ReLU activation\\
          
    &&3. $256$ kernels of size $4 \times 4$, stride size = $2$, ReLU activation. Followed by flattening \\
          
    &&4. Fully-connected layer with $768$ units, ReLU activation\\
   \hline
%%%%%%%%%%%%%%%%%%%%%%%%%%%%%%%%%%%%%%%%%%%%%%%%%%%%%%%%%%%%
    \multirow{4}{*}{CNN 2} & \multirow{4}{*}{4 layers }&1. $4$ kernels of size $4 \times 4$, stride size = $2$, leaky ReLU activation\\
      
    &&2. $8$ kernels of size $4 \times 4$, stride size = $2$, leaky ReLU activation\\
          
    &&3. $256$ kernels of size $4 \times 4$, stride size = $2$, ReLU activation. Followed by flattening \\
          
    &&4. Fully-connected layer with $768$ units, ReLU activation\\
   \hline
   %%%%%%%%%%%%%%%%%%%%%%%%%%%%%%%%%%%%%%%%%%%%%%%%%%%%%%%%%%%%
    \multirow{4}{*}{CNN 3} & \multirow{4}{*}{4 layers }&1. $4$ kernels of size $4 \times 4$, stride size = $2$, leaky ReLU activation\\
      
    &&2. $8$ kernels of size $4 \times 4$, stride size = $2$, leaky ReLU activation\\
          
    &&3. $32$ kernels of size $4 \times 4$, stride size = $2$, ReLU activation. Followed by flattening \\
          
    &&4. Fully-connected layer with $32$ units with linear activation as the mean, $\mu(\mathbf{j}_{1:t}^{\text{diff}})$.\\
   \hline
   %%%%%%%%%%%%%%%%%%%%%%%%%%%%%%%%%%%%%%%%%%%%%%%%%%%%%%%%%%%%
    \multirow{6}{*}{CNN 4} & \multirow{6}{*}{5 layers }&1. $4$ kernels of size $4 \times 4$, stride size = $2$, leaky ReLU activation\\
      
    &&2. $8$ kernels of size $4 \times 4$, stride size = $2$, leaky ReLU activation\\
          
    &&3. $16$ kernels of size $4 \times 4$, stride size = $2$, ReLU activation. Followed by flattening \\
          
    &&4. Fully-connected layer with $768$ units, linear activation\\
      
    &&5. Two branches of fully-connected layers with $32$ units each, as the mean and variance
    \\&& , $\mu_{\psi}$ and $\Sigma_{\psi}$, with linear activation\\
   \hline
%%%%%%%%%%%%%%%%%%%%%%%%%%%%%%%%%%%%%%%%%%%%%%%%%%%%%%%%%%%%
      \multirow{4}{*}{CNN 5} & \multirow{4}{*}{4 layers }&1. $4$ kernels of size $4 \times 4$, stride size = $2$, leaky ReLU activation\\
      
    &&2. $8$ kernels of size $4 \times 4$, stride size = $2$, leaky ReLU activation\\
          
    &&3. $16$ kernels of size $4 \times 4$, stride size = $2$, ReLU activation. Followed by flattening \\
          
    &&4. Fully-connected layer with $768$ units, linear activation\\
   \hline
%%%%%%%%%%%%%%%%%%%%%%%%%%%%%%%%%%%%%%%%%%%%%%%%%%%%%%%%%%%%
     \multirow{4}{*}{CNN 6} & \multirow{4}{*}{4 layers }&1. $4$ kernels of size $4 \times 4$, stride size = $2$, leaky ReLU activation\\
      
    &&2. $8$ kernels of size $4 \times 4$, stride size = $2$, leaky ReLU activation\\
          
    &&3. $16$ kernels of size $4 \times 4$, stride size = $2$, ReLU activation. Followed by flattening \\
          
    &&4. Fully-connected layer with $768$ units, linear activation\\
   \hline
%%%%%%%%%%%%%%%%%%%%%%%%%%%%%%%%%%%%%%%%%%%%%%%%%%%%%%%%%%%%
      MLP 2 & 1 layer & 1 .256 units, leaky ReLU activation\\
   \hline
%%%%%%%%%%%%%%%%%%%%%%%%%%%%%%%%%%%%%%%%%%%%%%%%%%%%%%%%%%%%
      \multirow{2}{*}{MLP 3} & \multirow{2}{*}{1 layer }&1. Two branches of fully-connected layers with $32$ units each, as the mean and variance
    \\&& , $\mu_{\phi}$ and $\Sigma_{\phi}$, with linear activation\\
   \hline
%%%%%%%%%%%%%%%%%%%%%%%%%%%%%%%%%%%%%%%%%%%%%%%%%%%%%%%%%%%%
      \multirow{6}{*}{CNN 7} & \multirow{6}{*}{5 layers }&1. Fully-connected layer with $768$ units, leaky ReLU activation\\
      
    &&2. Fully-connected layer with $6144$ units, leaky ReLU activation. Followed by reshaping to a\\
    && $(12, 2, 256)$ tensor\\
          
    &&3. Deconv. layer with $128$ kernels of size $5 \times 3$, stride size = $2$, leaky ReLU activation \\
          
    &&4. Deconv layer with $64$ kernels of size $6 \times 4$, stride size = $2$, leaky ReLU activation\\
              
    &&5. Deconv layer with $3$ kernels of size $3 \times 2$, stride size = $2$, sigmoid activation\\
\Xhline{2\arrayrulewidth}
%%%%%%%%%%%%%%%%%%%%%%%%%%%%%%%%%%%%%%%%%%%%%%%%%%%%%%%%%%%%   
  \caption{\small Detail of the prediction module for the NGSIM I-80 dataset. The coefficient for the leaky ReLU activation functions is $0.2$.}
  \label{table:ngsim}
  \end{longtable}
}
%\end{table*}
%%%%%%%%%%%%%%%%%%%%%%%%%%%%%%%%%%%%%%%%%%%%%%%%%%%%%%%%%
\normalsize
For difference learning model, we use the exact same structure of the networks. However, since the output of the module is the difference between two consecutive frames, it can get any values between -1 and 1. Therefore, we use \textit{tanh} activation for the last layer of network CNN 7.
\vspace{.5cm}
\subsection{Argoverse dataset}
The images in this dataset are $256 \times 256 \times 2$. They have two channels. One channel for OGM and the other one for the ego-vehicle. The ego-vehicle is  more towards the left side of the image. Therefore we first zero pad the left side of the image so that the ego-vehicle is centered and then zero pad the whole image to apply the functions in the IOTs. To extract the actions, we apply inverse of the functions in Fig. \ref{fig: measure} to compute the actions using the xy coordinate of the car at each time. Table \ref{table:argoverse} shows the details of the difference learning model, i.e. the model with the best performance for this dataset.  

\newpage
%%%%%%%%%%%%%%%%%%%%%%%%%%%%%%%%%%%%%%%%%%%%%%%%%%%%%%%%%
\footnotesize{
\begin{longtable}{l||l| L{10.5cm} }
\Xhline{1\arrayrulewidth}
	Network	& \multicolumn{2}{l}{Detail of structure} 	\\ 
\Xhline{2\arrayrulewidth}
%%%%% MLP 1 %%%%%%%%%%%%%%%%%%%%%%%%%%%%%%%%%%%%%%%%%%%%%%%%%%%%%%%
    \multirow{2}{*}{MLP 1} & \multirow{2}{*}{2 layers }& 1. $85$ units- leaky ReLU activation   \\
    
    && 2. $25$ units- ReLU activation\\
   \hline
%%%%% CNN 1 %%%%%%%%%%%%%%%%%%%%%%%%%%%%%%%%%%%%%%%%%%%%%%%%%%%%%%%
    \multirow{6}{*}{CNN 1} & \multirow{6}{*}{6 layers }&1. $16$ kernels of size $4 \times 4$, stride size = $2$, leaky ReLU activation\\
    
    &&2. $32$ kernels of size $4 \times 4$, stride size = $2$, leaky ReLU activation\\
        
    &&3. $64$ kernels of size $4 \times 4$, stride size = $2$, leaky ReLU activation\\
    
    &&4. $128$ kernels of size $4 \times 4$, stride size = $2$, leaky ReLU activation\\
    
    &&5. $256$ kernels of size $4 \times 4$, stride size = $2$, ReLU activation. Followed by flattening \\
       
    &&6. Fully-connected layer with $128$ units, ReLU activation\\
   \hline
%%%%% CNN 2 %%%%%%%%%%%%%%%%%%%%%%%%%%%%%%%%%%%%%%%%%%%%%%%%%%%%%%%
    \multirow{6}{*}{CNN 2} & \multirow{6}{*}{6 layers }&1. $2$ kernels of size $4 \times 4$, stride size = $2$, leaky ReLU activation\\
    
    &&2. $4$ kernels of size $4 \times 4$, stride size = $2$, leaky ReLU activation\\
        
    &&3. $8$ kernels of size $4 \times 4$, stride size = $2$, leaky ReLU activation\\
    
    &&4. $16$ kernels of size $4 \times 4$, stride size = $2$, leaky ReLU activation\\
   
    &&5. $256$ kernels of size $4 \times 4$, stride size = $2$, ReLU activation. Followed by flattening \\
       
    &&6. Fully-connected layer with $128$ units, ReLU activation\\
    \hline
    %%%%% CNN 3 %%%%%%%%%%%%%%%%%%%%%%%%%%%%%%%%%%%%%%%%%%%%%%%%%%%%%%%
        \multirow{6}{*}{CNN 3} & \multirow{6}{*}{6 layers }&1. $2$ kernels of size $4 \times 4$, stride size = $2$, leaky ReLU activation\\
    
    &&2. $4$ kernels of size $4 \times 4$, stride size = $2$, leaky ReLU activation\\
        
    &&3. $8$ kernels of size $4 \times 4$, stride size = $2$, leaky ReLU activation\\
    
    &&4. $16$ kernels of size $4 \times 4$, stride size = $2$, leaky ReLU activation\\
   
    &&5. $32$ kernels of size $4 \times 4$, stride size = $2$, ReLU activation. Followed by flattening \\
       
    &&6. Fully-connected layer with $32$ units with linear activation as the mean  , $\mu(\mathbf{j}_{1:t}^{\text{diff}})$.\\
    \hline
    %%%% CNN 4 %%%%%%%%%%%%%%%%%%%%%%%%%%%%%%%%%%%%%%%%%%%%%%%%%%%%%%%
    \multirow{8}{*}{CNN 4} & \multirow{8}{*}{7 layers }&1. $2$ kernels of size $4 \times 4$, stride size = $2$, leaky ReLU activation\\
  
    &&2. $4$ kernels of size $4 \times 4$, stride size = $2$, leaky ReLU activation\\
        
    &&3. $8$ kernels of size $4 \times 4$, stride size = $2$, leaky ReLU activation\\

    &&4. $16$ kernels of size $4 \times 4$, stride size = $2$, leaky ReLU activation\\

    &&5. $256$ kernels of size $4 \times 4$, stride size = $2$, ReLU activation. Followed by flattening \\

    &&6. Fully-connected layer with $128$ units, linear activation\\
  
    &&7. Two branches of fully-connected layers with $32$ units each, as the mean and variance
    \\&& , $\mu_{\psi}$ and $\Sigma_{\psi}$, with linear activation\\
   \hline
%%%%% CNN 5 %%%%%%%%%%%%%%%%%%%%%%%%%%%%%%%%%%%%%%%%%%%%%%%%%%%%%%%
    \multirow{6}{*}{CNN 5} & \multirow{6}{*}{6 layers }&1. $2$ kernels of size $4 \times 4$, stride size = $2$, leaky ReLU activation\\
  
    &&2. $4$ kernels of size $4 \times 4$, stride size = $2$, leaky ReLU activation\\
      
    &&3. $8$ kernels of size $4 \times 4$, stride size = $2$, leaky ReLU activation\\
  
    &&4. $16$ kernels of size $4 \times 4$, stride size = $2$, leaky ReLU activation\\

    &&5. $256$ kernels of size $4 \times 4$, stride size = $2$, ReLU activation. Followed by flattening \\
 
    &&6. Fully-connected layer with $128$ units, linear activation\\
   \hline
%%%%% CNN 6 %%%%%%%%%%%%%%%%%%%%%%%%%%%%%%%%%%%%%%%%%%%%%%%%%%%%%%%
    \multirow{6}{*}{CNN 6} & \multirow{6}{*}{6 layers }&1. $2$ kernels of size $4 \times 4$, stride size = $2$, leaky ReLU activation\\
 
    &&2. $4$ kernels of size $4 \times 4$, stride size = $2$, leaky ReLU activation\\
       
    &&3. $8$ kernels of size $4 \times 4$, stride size = $2$, leaky ReLU activation\\
 
    &&4. $16$ kernels of size $4 \times 4$, stride size = $2$, leaky ReLU activation\\
    
    &&5. $256$ kernels of size $4 \times 4$, stride size = $2$, ReLU activation. Followed by flattening \\

    &&6. Fully-connected layer with $128$ units, linear activation\\
   \hline
%%%%% MLP 2 %%%%%%%%%%%%%%%%%%%%%%%%%%%%%%%%%%%%%%%%%%%%%%%%%%%%%%%
      MLP 2 & 1 layer & 1 .256 units, leaky ReLU activation\\
   \hline
%%%%% MLP 3 %%%%%%%%%%%%%%%%%%%%%%%%%%%%%%%%%%%%%%%%%%%%%%%%%%%%%%%
      \multirow{2}{*}{MLP 3} & \multirow{2}{*}{1 layer }&1. Two branches of fully-connected layers with $32$ units each, as the mean and variance
    \\&& , $\mu_{\phi}$ and $\Sigma_{\phi}$, with linear activation\\
   \hline
%%%%% CNN 7 %%%%%%%%%%%%%%%%%%%%%%%%%%%%%%%%%%%%%%%%%%%%%%%%%%%%%%%
      \multirow{6}{*}{CNN 7} & \multirow{6}{*}{7 layers }&1. Fully-connected layer with $128$ units, leaky ReLU activation\\

    &&2. Fully-connected layer with $9216$ units, leaky ReLU activation. Followed by reshaping to a \\
    && $(6, 6, 256)$\\
        
    &&3. Deconv. layer with $128$ kernels of size $4 \times 4$, stride size = $2$, leaky ReLU activation. \\
      
    &&4. Deconv layer with $64$ kernels of size $4 \times 4$, stride size = $2$, leaky ReLU activation\\
            
    &&5. Deconv. layer with $32$ kernels of size $4 \times 4$, stride size = $2$, leaky ReLU activation. \\
       
    &&6. Deconv layer with $16$ kernels of size $4 \times 4$, stride size = $2$, leaky ReLU activation\\
         
    &&7. Deconv layer with $2$ kernels of size $3 \times 2$, stride size = $2$, tanh activation\\
\Xhline{2\arrayrulewidth}
  \caption{\small Detail of the prediction module for the Argoverse dataset as a part of difference learning module. The coefficient for the leaky ReLU activation functions is $0.2$. }
  \label{table:argoverse}
\end{longtable}
}
%%%%%%%%%%%%%%%%%%%%%%%%%%%%%%%%%%%%%%%%%%%%%%%%%%%%%%%%%

\end{document}